\documentclass[sigconf]{acmart}
\usepackage{amsmath}
\usepackage{amsfonts}
\usepackage{graphicx}
\usepackage{subcaption}
\usepackage{tabularx}
\usepackage{xcolor}
\usepackage{multirow}
\AtBeginDocument{%
  }
    
\setcopyright{acmcopyright}
\copyrightyear{2023}
\acmYear{2023}
\acmDOI{XXXXXXX.XXXXXXX}

\newcommand{\dtrain}{$\mathcal{D}^{train}$}

\begin{document}

\title{Understanding the Overfitting of the Episodic Meta-training}



% article.
% \author{Siqi Hui}
% \email{huisiqi@stu.xjtu.edu.cn}

% \author{Sanping Zhou}
% \email{spzhou@xjtu.edu.cn}

% \author{Ye Deng}
% \email{dengye@stu.xjtu.edu.cn}

% \author{Jinjun Wang}
% \email{jinjun@mail.xjtu.edu.cn}

% \affiliation{%
\author{Siqi Hui}
\email{huisiqi@stu.xjtu.edu.cn}
\affiliation{%
  \institution{Institute of Artificial Intelligence and Robotic of Xi'an Jiaotong University}
\city{Xi'an}
\country{China}
}

\author{Sanping Zhou}
\email{spzhou@xjtu.edu.cn}
\affiliation{%
  \institution{Institute of Artificial Intelligence and Robotic of Xi'an Jiaotong University}
\city{Xi'an}
\country{China}
}

\author{Ye Deng}
\email{dengye@stu.xjtu.edu.cn}
\affiliation{%
  \institution{Institute of Artificial Intelligence and Robotic of Xi'an Jiaotong University}
\city{Xi'an}
\country{China}
}

\author{Jinjun Wang}
\email{jinjun@mail.xjtu.edu.cn}
\affiliation{%
  \institution{Institute of Artificial Intelligence and Robotic of Xi'an Jiaotong University}
\city{Xi'an}
\country{China}
}

% \affiliation{%

% \affiliation{%

% \affiliation{%

% \affiliation{%

% \affiliation{%

% \affiliation{%

% \affiliation{%

% \affiliation{%
%   \institution{Institute of Artificial Intelligence and Robotics}
%   \city{Xian}
%   \state{ShaanXi}
%   \country{China}
%   \postcode{710049}
% }

\begin{abstract}
Despite the success of two-stage few-shot classification methods, in the episodic meta-training stage, the model suffers severe overfitting. We hypothesize that it is caused by over-discrimination, i.e., the model learns to over-rely on the superficial features that fit for base class discrimination while suppressing the novel class generalization. To penalize over-discrimination, we introduce knowledge distillation techniques to keep novel generalization knowledge from the teacher model during training. Specifically, we select the teacher model as the one with the best validation accuracy during meta-training and restrict the symmetric Kullback-Leibler (SKL) divergence between the output distribution of the linear classifier of the teacher model and that of the student model. This simple approach outperforms the standard meta-training process. We further propose the Nearest Neighbor Symmetric Kullback-Leibler (NNSKL) divergence for meta-training to push the limits of knowledge distillation techniques. NNSKL takes few-shot tasks as input and penalizes the output of the nearest neighbor classifier, which possesses an impact on the relationships between query embedding and support centers. By combining SKL and NNSKL in meta-training, the model achieves even better performance and surpasses state-of-the-art results on several benchmarks.
\end{abstract}




\maketitle
\section{Introduction}
Humans possess exceptional learning abilities that enable them to generalize from a small number of examples to new situations. Nonetheless, current deep-learning techniques rely heavily on vast amounts of training data. Consequently, few-shot learning was introduced to teach networks how to understand new concepts with only a few labeled examples, and transfer learning has emerged as a practical alternative to directly learning large numbers of parameters with limited samples. This approach involves training networks on common or base classes with sufficient samples and transferring the model to learn novel classes with a limited number of examples.

\begin{figure}[thbp]
  \centering
    \includegraphics[width=0.95\linewidth]{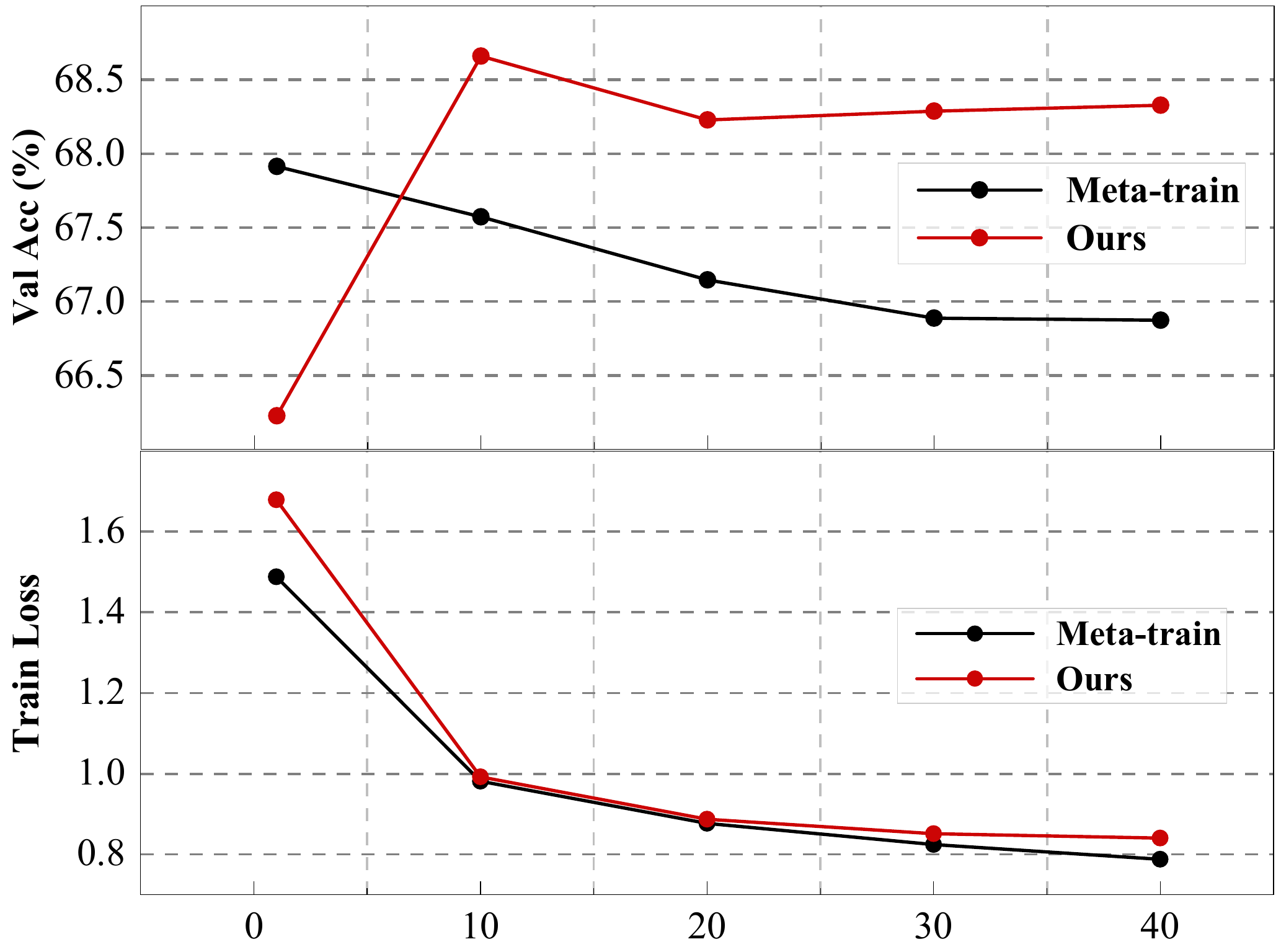}
  \caption{Validation accuracy of 1-shot tasks and supervised contrastive loss on the CIFAR-FS dataset \cite{bertinetto2018meta}. ResNet12 is chosen as the backbone network and was meta-trained forty epochs to perform 1-shot few-shot tasks on base classes. The standard meta-train model was trained with only supervised contrastive loss (black lines) while our model is trained by the combination of SKL, NNSKL and supervised contrastive loss (red lines).}
  % \caption{The training loss curves (orange) and validation accuracy curves (blue) in the meta-training stage.}
  \label{fig:overdis}
\vspace{-0.5cm}
\end{figure}

The meta-learning framework is a representative few-shot learning approach that leverages the concept of learning to learn \cite{nichol2018reptile, finn2017model, snell2017prototypical}. This framework selects few-shot classification tasks or episodes from the training samples that belong to the base classes and optimizes the model to achieve high performance on these tasks. A typical task involves an \emph{N}-way and \emph{K}-shot scenario, where \emph{N} classes are represented by \emph{K} support samples and \emph{Q} query samples in each class. The objective is to classify the $N \times Q$ query samples into the \emph{N} classes based on the $N\times K$ support samples. During the meta-testing phase, the model is evaluated on a series of tasks similar to the meta-training tasks. Optimization-based meta-learning methods differentiate the model through the optimization process over a support set to generate a model that can be adapted to novel tasks using a few optimization steps \cite{nichol2018reptile, finn2017model}. On the other hand, metric-based methods meta-learn deep metric spaces for comparing query and support embeddings, which can be directly adapted to solve meta-testing tasks without task-specific optimization \cite{snell2017prototypical, vinyals2016matching, sung2018learning}. By episodically optimizing the model with supervised contrastive loss \cite{snell2017prototypical}, metric-based methods learn the representations with good \emph{base class discrimination} \cite{chen2021meta}. These methods perform well when the novel classes and base classes share modes of variation, class-distinctive features, or other inductive biases \cite{phoo2020self}. Nevertheless, they generalize poorly when novel classes exhibit extremely different data distributions \cite{Triantafillou2019}.

 Compared to the meta-learning framework, simple transfer learning (STL) \cite{Dhillon2019, chen2019closer, Triantafillou2019, Tian2020, bendou2022easy, Xu2023} has been found to outperform meta-learning algorithms. The STL framework involves training a deep neural network non-episodically to perform whole-classification, using the standard cross-entropy loss on the entire base classes, which results in embeddings exhibiting stronger \emph{novel class generalization}. In other words, the model performs better on tasks of novel classes that are more dissimilar from the base classes. For each meta-testing task, current methods either apply the nearest neighbor classifier or finetune a new linear classifier on top of the trained feature extractor to discriminate between unseen classes.

In recent years, the two-stage training paradigm has gained popularity in few-shot learning, where non-episodic pre-training is followed by episodic meta-training to achieve state-of-the-art performance \cite{Liu2021,ye2020few,rusu2018meta, Shen2020, Wertheimer2021, Rusu2019, chen2021meta}. During pre-training, the model is trained to perform supervised learning or a combination of supervised learning and self-supervised learning on the base classes, which provides a base learner with good novel class generalization for the meta-training stage \cite{chen2021pareto,su2020does,gidaris2019boosting,bendou2022easy, Liu2021, 2019When}. In the meta-training stage, the model is episodically trained by the supervised contrastive loss, which boosts the \emph{base class discrimination}. Despite the success of the meta-training stage \cite{Jian2022, Hu2022, Wertheimer2021, chen2021meta}, the model is susceptible to overfitting (see Figure \ref{fig:overdis}). This occurs when the model shows less discrimination for novel classes, and the validation accuracy decreases along with the decrease of the supervised contrastive loss on the meta-training dataset. We hypothesize that this is caused by over-discrimination, i.e., the model relies too heavily on the superficial features that are useful for \emph{base class discrimination} while neglecting the knowledge that exhibits robust \emph{novel class generalization}, as there are only supervision signals from the supervised contrastive loss on the base class and novel class generalization is not paid any attention. Without balancing \emph{novel class generalization} and \emph{base class discrimination} during meta-training, current methods indirectly design schedulers to gradually lower the learning rate and rely on cross-validation to select the model that generalizes best on the novel class, which results in sub-optimal solutions. 

In this paper, to strike a good balance between the \emph{novel class generalization} and the \emph{base class discrimination}, we introduce knowledge distillation techniques that preserve generalizable knowledge during training. We also propose a novel meta-training process for the two-stage few-shot framework. Specifically, we choose the model with the best validation accuracy as the teacher model and penalize the standard symmetric KL divergence (SKL) between the output distribution of the fully connected layer of the teacher model and that of the student model, in addition to the supervised contrastive loss (see Figure \ref{fig:metatrain} (a)). SKL helps to consider the relationships between the student and teacher embeddings, which mitigates the over-discrimination problem and boosts few-shot performances. To further improve the ability of knowledge distillation techniques, we propose the Nearest Neighbor Symmetric KL divergence (NNSKL) for few-shot classification. In contrast to SKL, which considers individual images, NNSKL takes few-shot tasks as input and penalizes the symmetric KL divergence between the output distributions of the nearest-neighbor classifier of the teacher and the student model. NNSKL aims to penalize the relationships between query and support embeddings and can be seamlessly combined with the supervised contrastive loss. As a result, NNSKL is better suited for meta-training and leads to improved performance. By combining SKL and NNSKL, we observe even better few-shot performance.

To systematically evaluate the effectiveness of our proposed approach, we conduct experiments on four benchmark datasets, namely  miniImageNet \cite{cai2018memory}, tieredImageNet \cite{ren2018meta}, CIFAR-FS \cite{bertinetto2018meta}, and CUB \cite{wah2011caltech}. Our results show that the student model trained with standard symmetric knowledge distillation loss and NNSKL outperforms the baseline model meta-trained with only supervised contrastive loss. Our approaches also achieve state-of-the-art performances on all four datasets. We also devise ablative studies to study the effectiveness of knowledge distillation techniques for mitigating overfitting in the meta-training stage for few-shot image classification.

In summary, our contributions are three-folded:
\begin{itemize}
    \item We identify the issue of severe overfitting during the meta-training stage of the two-stage training framework, caused by over-discrimination on base classes.
    \item We propose SKL and NNSKL to preserve the generalizable features and the relationships between query and support embeddings. 
    \item We propose a novel meta-training process that incorporates SKL and NNSKL for few-shot classification, which mitigates the overfitting and achieves state-of-the-art classification performance on popular few-shot learning datasets.
\end{itemize}

\section{Related Works}
\subsection{Learning Frameworks for FSL}
Few-shot learning is a challenging task that enables models to quickly adapted to classify images of unseen classes, given limited labeled samples. Few-shot classification has received significant attention in recent years, and various approaches have been proposed to address this problem. Based on the training framework, current few-shot learning methods can be roughly grouped into three categories of meta-learning, simple transfer learning, and two-stage learning. 

One of the representative few-shot learning frameworks is meta-learning. In meta-learning, the model is trained to perform a series of meta-training tasks (episodes) which are composed of base class images and are then tested on meta-testing tasks with disjoint classes. These meta-learning architectures can be broadly categorized into two groups. Optimization-based methods \cite{grant2018recasting, rusu2018meta} differentiate an optimization process over the support set within the meta-learning framework and learn a model that adapted to novel tasks through steps of the optimization process. For instance, MAML \cite{finn2017model} identifies a neural network initialization that can be adapted to any novel task using a few optimization steps. MetaOptNet \cite{lee2019meta} learns a feature representation that generalizes well for a linear support vector machine (SVM) classifier. In contrast, metric-based methods \cite{vinyals2016matching} meta-learn a deep representation with a metric in feature space. For example, Prototypical Networks \cite{snell2017prototypical} compute the average feature for each class in the support set and classify query samples using the nearest-centroid method with the Euclidean distance as a Bregman divergence. Relation Networks \cite{sung2018learning} propose a relation module as a learnable metric jointly trained with deep representations to generalize this framework.

Rather than meta-learning on a series of episodes, it has been found that the STL framework, which does not rely on episodic training at all, has been shown to achieve competitive performance. The STL framework involves training a deep neural network on the base classes using standard cross-entropy loss \cite{chen2019closer,dhillon2019baseline,tian2020rethinking}. For instance, Coseine Classifier \cite{gidaris2018dynamic} and Baseline++ \cite{chen2019closer} use whole-classification training by replacing the top linear layer with a cosine classifier. They then adapt the classifier to a few-shot classification task of novel classes by performing the nearest centroid or fine-tuning a new layer on top of the feature extractor, respectively. These works demonstrate that whole-classification models can achieve competitive or even better performances compared to several popular meta-learning models. Some other methods also use self-supervised tasks \cite{chen2021pareto,su2020does,gidaris2019boosting,bendou2022easy} (i.e., loss of contrastive learning \cite{Liu2021}, rotation prediction \cite{2019Boosting}, relative patch location \cite{2019Boosting} or image jigsaw puzzle \cite{2019When}) as auxiliary tasks to further learning embeddings with novel class generalization.

Recently, the two-stage learning framework of pre-training followed by meta-training is commonly used to achieve state-of-the-art few-shot performances \cite{Bateni2020, Ye2020, Liu2021, yang2022few,guo2022learning}. In the pre-training stage, same as STL methods, the model is trained to perform the whole-classification and provides features that exhibit novel class generalization for the following meta-training stage. Then, taken as the base learner, the pre-trained model is episodically meta-trained to perform supervised contrastive learning on base classes, which learns a deep metric space that fits for comparing embeddings in the low-shot regime. The meta-training stage is known to significantly boost the base class discrimination \cite{chen2021meta} and results in better few-shot classification performances. Meta-baseline \cite{chen2021meta} meta-learns over a whole-classification pre-trained model. Some recent dense-feature-based meta-learning methods \cite{Liu2021, Wertheimer2021} pre-train models based on the classification layer parameterized by a set of class-specific feature maps.
\vspace{-0.1cm}

\subsection{Knowledge Distillation for FSL}
 Inspired by the way human beings learn, there has been a long line of research and development on transferring knowledge from one model to another. Breiman and Shang \cite{breiman1996born} first proposed learning singletree models that approximate the performance of multiple tree models and provide better interpretability. Similar approaches for neural networks have emerged in the
work of Bucilua et al. \cite{buciluǎ2006model}, Ba and Caruana \cite{ba2014deep}, and Hinton et al. \cite{hinton2015distilling}, mainly for the purpose of model compression. A few recent papers have shown that distilling a teacher model into a student model of identical architecture, i.e., self-distillation, can improve the student over the teacher. Furlanello et al. \cite{furlanello2018born} and Bagherinezhad et al. \cite{bagherinezhad2018label} demonstrate it by training the student using softmax outputs of the teacher as ground truth over generations. Yim
et al. \cite{yim2017gift} transfer output activations using Gramian matrices and then fine-tune the student. 

In recent years, the STL framework has employed knowledge distillation techniques to enhance embeddings for meta-testing, resulting in superior \emph{novel class generalization} \cite{Jian2022}. Tian et al. \cite{tian2020rethinking} introduced a straightforward baseline that surpassed meta-learning algorithms. The approach involves first training a representation model through whole-classification and then refining the representations using self-distillation in the second stage. Similarly, Rajasegaran et al. \cite{rajasegaran2020self} pre-train a teacher model and then distill the generalizable manifold knowledge from the teacher model while reducing the intra-class variance. Liu et al. \cite{liu2021few} proposed a unified one-stage method that incorporates both supervised training and knowledge distillation. Rizve et al. \cite{rizve2021exploring} employed self-distillation to underscore equivariance and invariance knowledge in the model.

Although knowledge distillation techniques have been studied by previous few-shot learning methods, they have only been introduced by the STL framework to improve the model's generalization for novel classes. No studies have applied knowledge distillation to mitigate the over-discrimination problem of the meta-training stage's two-stage training paradigm. To fill this gap, we introduce knowledge distillation techniques to preserve novel class generalization knowledge. To the best of our knowledge, we are the first to use knowledge distillation to combat over-discrimination caused by the supervised contrastive loss in the meta-training stage.

\begin{figure}[t]
\centering
\includegraphics[width=0.92\linewidth]{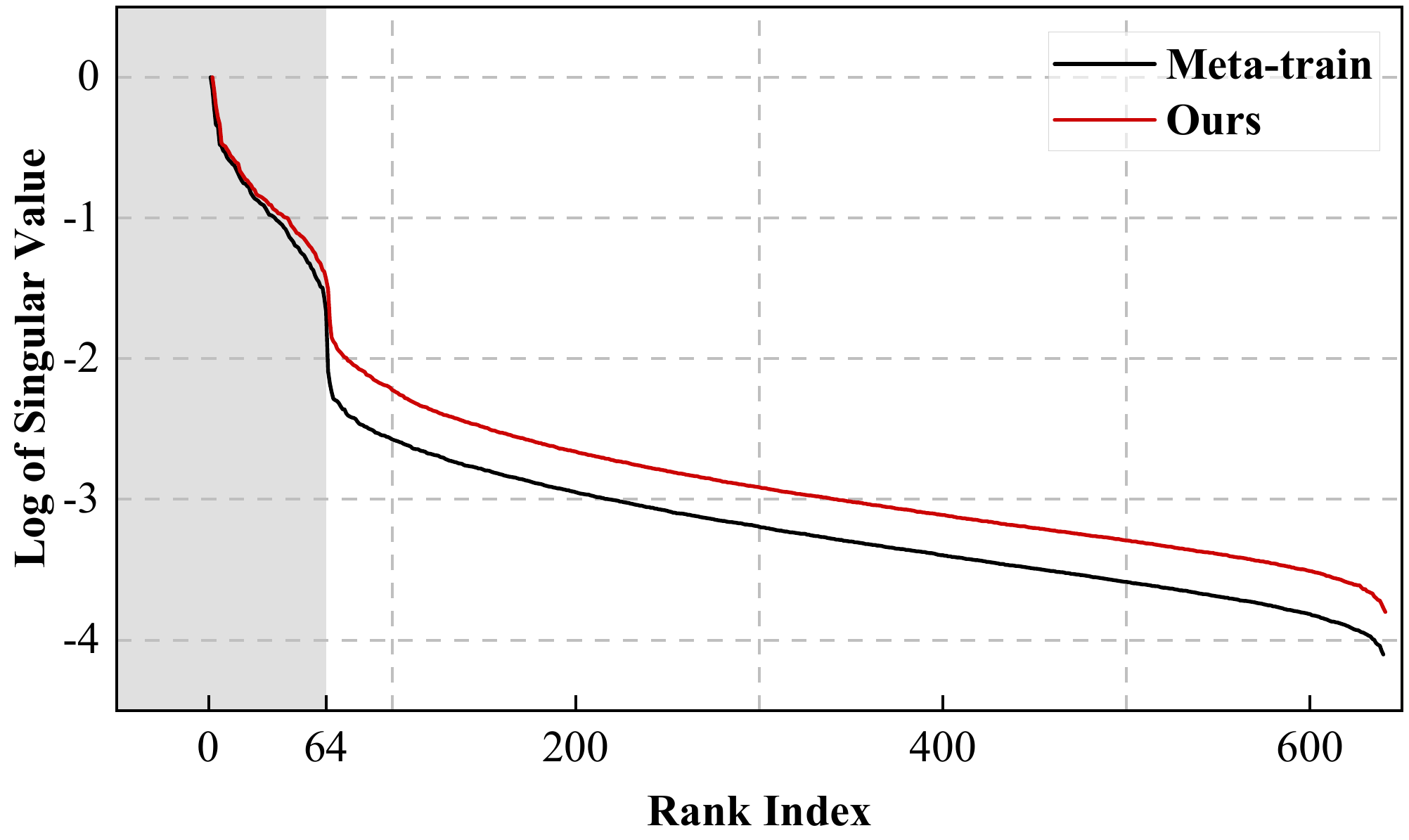}
\caption{Singular value spectrums of embedding matrices on the training set of CIFAR-FS. Embeddings are extracted by models trained 40 epochs to perform 5-shot tasks. Each spectrum contains 640 singular values in sorted order and logarithmic scale. Spectrums drop after the rank of 64, indicating dimension collapse.}
\label{fig:collapse}
\vspace{-0.3cm}
\end{figure}

\section{Method}
We present an overview of few-shot classification in Section \ref{preliminary}, followed by an analysis of the overfitting problem of meta-training in Section \ref{analysis}. In Section \ref{kd}, we discuss how knowledge distillation techniques can help mitigate the issue of overfitting. Finally, in Section \ref{sec:ourframework}, we propose our novel meta-training process.

\subsection{Preliminary}
\label{preliminary}
Few-shot classification is a type of learning that involves training a model capable of adapting quickly to new classification tasks with limited samples. During training, a training dataset $\mathcal{D}^{train}=\{(x_n,y_n)\}_{n=1}^{|\mathcal{D}^{train}|}$ is provided with $N_{train}$ classes, where $x_i \in \mathbb{R}^D$ represents the $i$-th image and $y_i\in [N_{train}]$ is its corresponding label. We can learn a model $f_\theta$ through a training framework. In the testing phase, a series of tasks or episodes $\{t_m\}_{m=1}^{|t|}$ are constructed from the test dataset $\mathcal{D}_{test}$ that consists of classes different from the training dataset. Each task $t$ contains a support set $\mathcal{S}={(x_j,y_j)}_{j=1}^{|\mathcal{S}|}$ and a query set $\mathcal{Q}={(x_i,y_i)}_{i=1}^{|\mathcal{Q}|}$, which are used to evaluate the trained model. The support set and the query set share the same label space. A task $t$ can be defined as an $N$-way-$K$-shot task if the support set has $N$ classes and each class has $K$ samples. To solve the task $t$, we can produce a nearest neighbor classifier $g(\cdot;f_\theta,\mathcal{S}):\mathbb{R}^D \mapsto [N]$ based on the nearest-centroid metric and the support set embeddings extracted by the learned model $f_\theta$. The performance is measured by the average classification accuracy over all sampled tasks.

\subsubsection{Pre-training.}
The two-stage learning paradigm is a popular training framework to solve the FSL task, which comprises a pre-training stage and a meta-training stage. In the pre-training stage, the model is trained non-episodically to perform whole-classification, thereby learning generalizable embeddings that are applicable to any-way and any-shot few-shot problems. To regularize these parameters, a regularization term is added, resulting in a total learning objective for the pre-training stage. The model in the pre-training stage is trained by the following learning objective:
\begin{equation}
    \mathcal{L}_{pre}=\underset{( x,y) \sim \mathcal{D}^{train}}{\mathbb{E}}\mathcal{L}_{ce}( g_{\phi}\circ f_{\theta}( x ),y ) +\mathcal{R}(\theta,\phi),
\end{equation}
where $g_{\phi}\left( X \right) =\mathbf{W}_{\phi}X+\mathbf{b}$ is the linear classification layer on the top of the feature extractor $f_\theta$, which provides logits for cross-entropy loss.

\subsubsection{Meta-training.}
In the episodic meta-training stage, the generalizable feature extractor from the pre-training stage is utilized as a base learner to learn a metric space that is better suited for meta-testing. Tasks are composed in an episodic manner to mimic true test tasks based on samplers from \dtrain, following the setting of \emph{N}-way-\emph{K}-shot task, where each task contains a support set and a query set. The output features of the query set and the support set are denoted as $f^q$, and $f^s$, respectively. The meta-training stage introduces supervised contrastive loss, similar to ProtoNet \cite{snell2017prototypical}, to stress the discriminative features and learn a metric space to compare embeddings. The supervised contrastive loss takes the following form:
\begin{equation}
\label{scl}
\begin{aligned}
\mathcal{L}_{SC}= \frac{1}{|\mathcal{Q}|}\sum_{i=1}^{|\mathcal{Q}|}{-\log\text{\ }p_{\theta}( y_{i}=k|x_i )},&\\
p_{\theta}(y_{i}=k|x_i) =\frac{ \exp ( -d( f_{i}^{q},c_k ) / \tau )}{ \sum_{k'\in [N]} 
 \exp ( -d(f_{i}^{q},c_{k'})/\tau)},&
\end{aligned}
\end{equation}

\noindent where $\tau$ is the temperature hyperparameter, which we set as 1 for all experiments for simplicity and clarity, and \emph{d} is the distance metric and we use the Euclidean distance in this paper, i.e. $d(f^q_i,c_{k}) =||f^q_i-c_{k}||$. Each class prototype $c_k=\frac{1}{K}\sum_{y_i=k} f^s_i$ is the mean vector of the embedding support points that belonging to class \emph{k}.

\subsection{Over-discrimination}
\label{analysis}
% During the meta-training stage, the model is trained using the supervised contrastive loss to enhance the metric space for measuring the distance between query embeddings and class prototypes. Although the supervised contrastive loss has shown to improve the model's performance, it can also result in severe overfitting, as depicted in the figure. We propose that this may be due to over-discrimination, whereby the model places excessive emphasis on superficial features that are relevant to the training tasks but overlooks crucial patterns that facilitate novel class generalization.

During meta-training, the model is trained using supervised contrastive loss to improve the metric space, which results in better performance when computing the distance between query embeddings and class prototypes. However, the supervised contrastive loss leads to severe overfitting (see Figure \ref{fig:overdis}), where the performance on the novel classes drops along with the increase of the base class performance. Our hypothesis is that this is caused by \emph{over-discrimination} on the base classes, where the model overemphasizes superficial features that are useful for base class discrimination but forgets essential patterns that are necessary for novel class generalization. To validate our hypothesis, we analyze the data distribution of the base classes as well as the gradient of the supervised contrastive loss.

\begin{figure}[th] 
  \centering
    \includegraphics[width=0.9\linewidth]{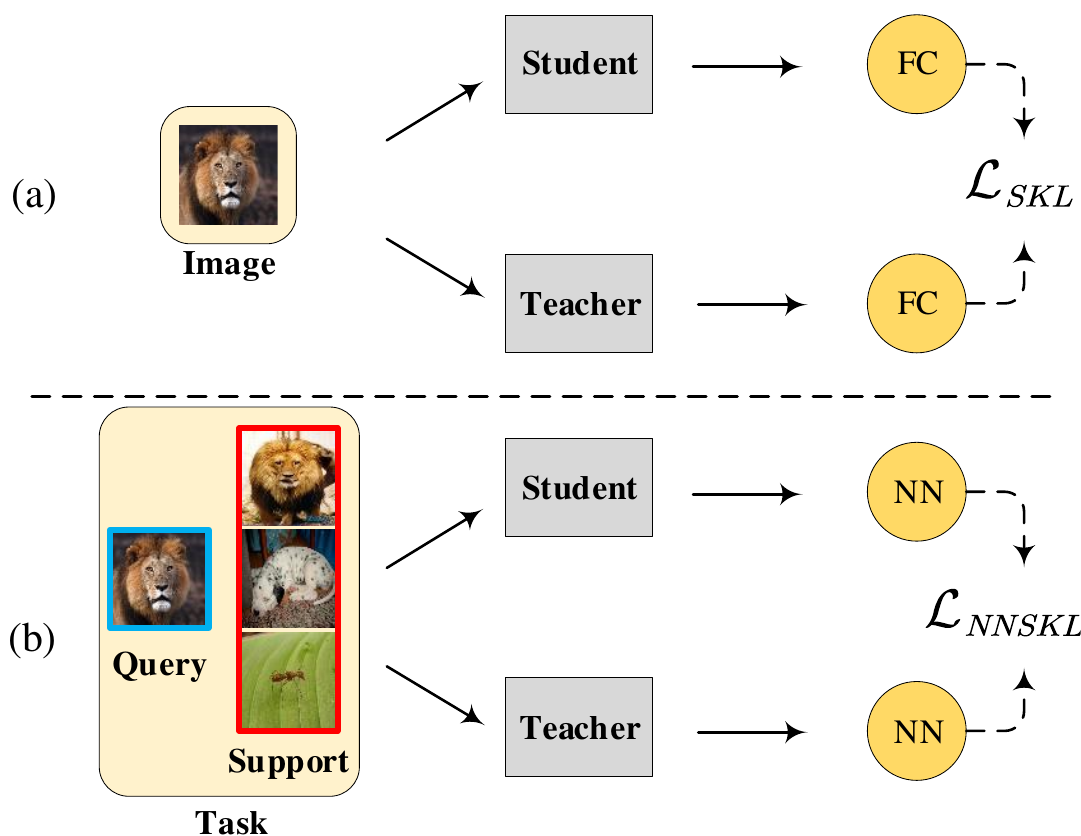}
    \caption{Visual intuition of SKL and NNSKL. \textbf{(a)} SKL takes individual images as input and penalizes the symmetric KL divergence between the output distribution of the linear classifier of the student and the teacher networks. \textbf{(b)} NNSKL takes few-shot tasks as input and penalizes the symmetric KL divergence between the output distribution of the nearest neighbor classifier of the teacher and the student models.}
    \label{fig:metatrain}
\vspace{-0.5cm}
\end{figure}

\subsubsection{Dimension Collapse}
During meta-training, over-discrimination causes severe dimensional collapse on the base classes, where the embedding vectors occupy a low dimensional subspace of the full dimensional embedding space \cite{Jing2021}. To illustrate the presence of the dimensional collapse, we conducted a singular value decomposition on the embedding matrix of the meta-training set of the CIFAR-FS, and we report the sorted, logarithmically scaled singular values in Figure \ref{fig:collapse}. As can be seen, at the rank index of 64, the sorted singular values show a substantial decline, indicating dimension collapse. It is worth noting that, the number of unbroken dimensions exactly matches the number of the base classes (64 for the CIFAR-FS dataset), which implies that the remaining dimensions contain information only enough to discriminate base classes while forgetting other information for novel class generalization.

\begin{figure*}[htbp]
\centering
    \includegraphics[width=\linewidth]{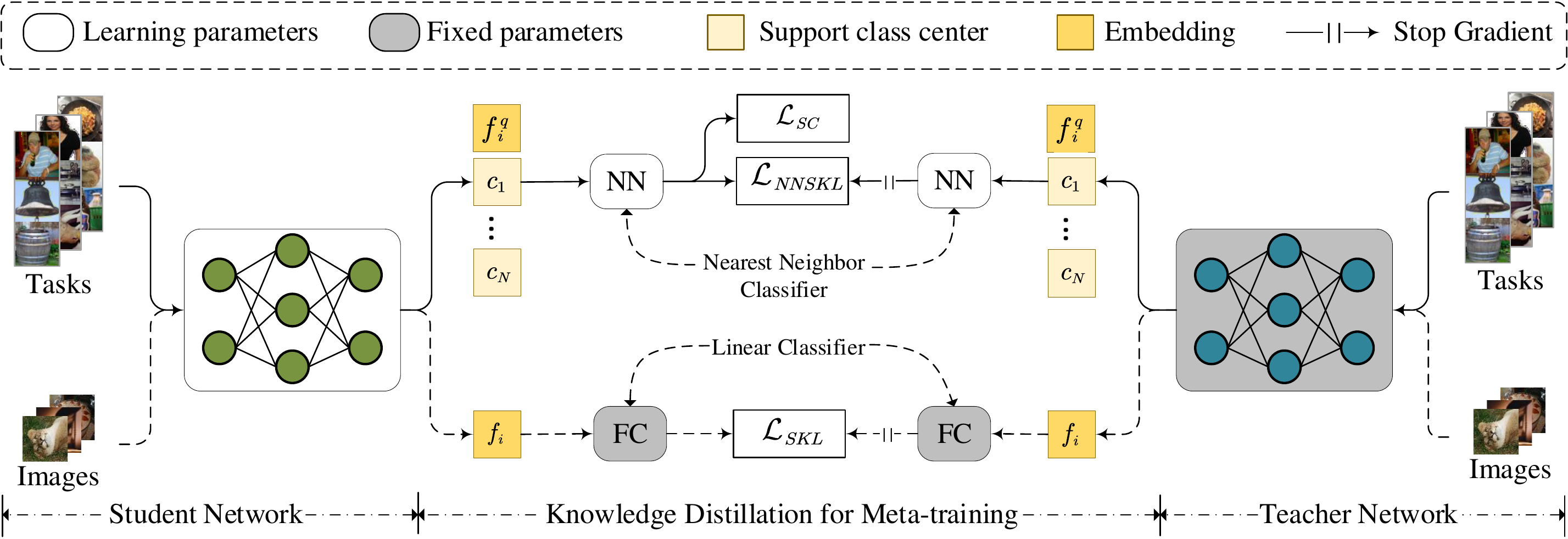}
    \caption{The meta-training process with SKL and NNSKL. Few-shot tasks and non-episodic images are first input to the student and teacher models to extract embeddings. Then, the query embeddings and support class centers are sent to nearest neighbor classifiers to compute NNSKL and the supervised contrastive loss. Meanwhile, the embeddings of non-episodic images are sent to linear classifiers to compute SKL.}
\label{fig:framework}
\end{figure*}

\subsubsection{Gradient Analysis} To gain a better understanding of the over-discrimination problem, we conducted a gradient analysis of the supervised contrastive loss with respect to each query embedding. Our findings show that the sparsity of the support set is the source of the \emph{over-discrimination}. Based on the Eq. \eqref{scl}, when given a single sample $x_i$, the gradient of $\mathcal{L}_{SC}$ with respect to $f^q_i$ is computed as:

\begin{equation}
\label{grad}
\begin{aligned}
    \frac{\partial \mathcal{L}_{SC}}{\partial f_{i}^{q}}=\sum_{k=1}^N&{\omega_{k}\Delta _{k}},\\
    \omega_{k}=\Big\{
    \begin{array}{c}
        \text{1}-p_{\theta}( k|x_i ) ,\ if\ y_i=k\\
        -p_{\theta}( k|x_i ) , \hspace{0.25cm}\ if\ y_i\ne k\\
    \end{array}&, \hspace{0.1cm}
    \Delta_{k}=\frac{f_{i}^{q}-c_{k}}{||f_{i}^{q}-c_{k}||}.
\end{aligned}
\end{equation}

 This gradient indicates that, after one step update, query embeddings would be moved towards the low-dimensional vector space generated by the set of few class prototypes. Additionally, due to the low-shot and low-way support dataset, the variance of the estimated prototypes is high, which could mislead the query embeddings. As a consequence, the model learns to over-rely on superficial features that help solve tasks on base classes while ignoring the features that better generalize to novel classes.

In this paper, we propose methods for improving base class discrimination while maintaining novel class generalization by distillation knowledge from a teacher model.

\subsection{Solving Over-discrimination}
\label{kd}

\subsubsection{Standard Symmetric KL Divergence}
 First, we utilize standard symmetric KL divergence \cite{aghajanyan2020better} as an auxiliary loss to mitigate over-discrimination. SKL takes batches of images as input and penalizes the output distribution of the final linear classifier of the student model and the teacher model. Suppose the teacher model can be parameterized as $f_{\theta'}$, the loss function is as follows:
\begin{equation}
\label{skl}
    \mathcal{L}_{SKL}=\underset{( x,y ) \sim \mathcal{D}_{train}}{\mathbb{E}}KL_S ( g_{\phi}\circ f_{\theta}( x ) ,g_{\phi}\circ f_{\theta'}( x )),
\end{equation}
where the $\theta'$ is the parameters of the teacher model and $KL_S$ is the symmetric KL divergence ($KL_S(X,Y)=KL(X||Y)+KL(Y||X)$) \cite{aghajanyan2020better, jiang2019smart}. According to Eq. \ref{skl}, given an image $x$, SKL introduces soft class distribution from the teacher model to prevent the output distribution of the student model from collapsing to one class, which implicitly restricts embeddings of the student model not far from the corresponding teacher embeddings \cite{kim2021comparing}, thus combating over-discrimination. The choice of the teacher model will be detailed in Section \ref{sec:ourframework}. 

\subsubsection{Nearest Neighbor KL Divergence} 
Second, to make full use of the generalization knowledge of the teacher model, we further propose NNSKL for the meta-training process. NNSKL takes batch of few-shot tasks as input and integrates supervisory signals into the nearest neighbor distributions used to construct the supervised contrastive loss. The NNSKL takes the following form:
\begin{equation}
    \mathcal{L}_{NNSKL}=\frac{1}{|\mathcal{Q}|}\sum_{i=1}^{|\mathcal{Q}|}{KL_S( p_{\theta}( x ) ,p_{\theta '}( x ) )}.
\end{equation}
Compared to SKL, NNSKL takes few-shot tasks as input which considers relationships between query and support embeddings, thus showing stronger penalization (see Figure \ref{fig:metatrain}). Besides, NNSKL can be seamlessly integrated into standard meta-training, as the query and support embeddings used by NNSKL can be also used to compose supervised contrastive loss. 

% \noindent\textbf{Gradient Analysis.} 
% To understand how SKL and NNSKL solve over-discrimination, we compute the gradient of SKL and NNSKL with respect to query embeddings. Defining the set of support embeddings and query embeddings extracted by the teacher model as $\hat{f^q}$ and $\hat{f^s}$, the gradient of $\mathcal{L}_{SKL}$ with respect to each embedding $f^q_i$ can be approximated as:
% \begin{equation}
% \label{gradskl}
% \begin{aligned}
%     &\frac{\partial \mathcal{L}_{SKL}}{\partial f_{i}^{q}}=\omega ^T\mathbf{W},\\
%     &\omega =\frac{1}{N_{train}}\left( f_{i}^{q}-\hat{f_{i}^{q}} \right) ,\\
%     &\mathbf{W}=\mathbf{W}_{\phi}^{T}\mathbf{W}_{\phi},
% \end{aligned}
% \end{equation}
% where $\hat{f^q_i}$ query embedding vector of the same query image provided by the teacher network, and $\mathbf{W}$ is a symmetric matrix with rank equals the rank of $\mathbf{W}_\phi$ ($N_{train}$ if full rank). Based on the above equation, gradients for query embeddings belong to a subspace spanned by vectors of $\mathbf{W}$. As the max rank of $\mathbf{W}$ is much larger than the rank of the support class number ($N_{train} \gg N$), SKL prevents embeddings from collapsing to low dimensional subspace, thus mitigating over-discrimination. 

Similarly, we also compute the approximated gradient of $\mathcal{L}_{NNSKL}$ with respect to an embedding $f^q_i$ as follows:
\begin{align}
\label{gradnnskl}
&\frac{\partial \mathcal{L}_{NNSKL}}{\partial f_{i}^{q}}=\sum_{k=1}^N{r_k\Delta _k},\\
\label{intraclass}
&r_k=\frac{1}{K}\left( ||f_{i}^{q}-c_k||-||f_{i}^{q}-\hat{c_k}|| \right),
\end{align}
where the $\frac{1}{K}\sum_{y_i=k}{\hat{f_{i}^{s}}}$ is the average of the teacher embeddings belonging to the \emph{k-th} support class. In Eq. \ref{intraclass} and Eq. \ref{gradnnskl}, for a class \emph{k}, when $||f_{i}^{q}-c_k||-||f_{i}^{q}-\hat{c_k}|| > 0$, NNSKL would shrink the distance between the query embedding $f^q_i$ and the class center $c_k$ (i.e., decrease $||f_{i}^{q}-c_k||$) and vice versa. In that, NNSKL addresses over-discrimination through directly optimizing intra-class variances and inter-class distances.

\subsection{Novel Meta-training Process.}
\label{sec:ourframework}
To take full advantage of knowledge distillation techniques, we propose a novel meta-training process that utilizes both SKL and NNSKL to boost the base class discrimination while preserving the novel class generalization of the teacher model. The student model is meta-learned through the following loss function: 
\begin{equation}
    \mathcal{L}_{meta}=\mathcal{L}_{SC}+\lambda_1 \mathcal{L}_{SKL} + \lambda_2 \mathcal{L}_{NNSKL},
\end{equation}
where $\lambda_1 and \lambda_2$ are loss weights for SKL and NNSKL. If not specified, we set $\lambda_1=\lambda_2=1$ for experiments on all datasets. At the start of meta-training, the pre-trained model is selected as the teacher model. Then, at the end of each epoch, if the validation accuracy of the student model is larger than that of the teacher model, the student model is selected as the teacher model for the next meta-training epoch. The whole framework of our meta-training process is shown in Figure \ref{fig:framework}. 

\section{Experiments}
In this section, we first provide a brief overview of the experimental settings. Then, we compare our results  with those of several state-of-the-art methods on four popular few-shot classification benchmarks, namely miniImageNet, tieredImageNet, CIFAR-FS, and CUB. Finally, we present ablation experiments and analyze how knowledge distillation impacts the deep metric space.

\begin{table*}[thbp]
\centering
\caption{Few-shot classification accuracy (\%) with 95\% confidence intervals on miniImageNet and tieredImageNet.}
\begin{tabularx}{\textwidth}{>{\centering\arraybackslash}X>{\centering\arraybackslash}X>{\centering\arraybackslash}X>{\centering\arraybackslash}X>{\centering\arraybackslash}X>{\centering\arraybackslash}X}
\hline
\multirow{2}{*}{Method} & \multirow{2}{*}{Backbone} & \multicolumn{2}{c}{\textbf{miniImageNet}} & \multicolumn{2}{c}{\textbf{tieredImageNet}} \\ \cline{3-6} 
                        &                           & 1shot          & 5-shot          & 1-shot           & 5-shot          \\ \hline
CTM  (2019)           & ResNet18          & 64.12 ± 0.82    & 80.51 ± 0.13            & 64.78 ± 0.11    & 81.05 ± 0.52               \\
FEAT (2020)           & ResNet12          & 66.78 ± 0.20    & 82.05 ± 0.14            & 70.80 ± 0.23    & 84.79 ± 0.16               \\
DeepEMD (2020)           & ResNet12          & 65.91 ± 0.82    & 82.41 ± 0.56            & 71.16 ± 0.87    & 86.03 ± 0.58               \\
S2M2  (2020)         & ResNet34          & 63.74 ± 0.18    & 79.45 ± 0.12            & -               & -                          \\
BML  (2021)           & ResNet12                  & 67.04 ± 0.63    & 83.63 ± 0.29            & 68.99 ± 0.50    & 85.49 ± 0.34               \\

InvEq  (2021)  &   ResNet12                & 67.28 ± 0.80    & 84.78 ± 0.52            & 72.21 ± 0.90    & 87.08 ± 0.58               \\
ReNet (2021)          & ResNet12          & 67.60 ± 0.44    & 82.58 ± 0.30            & 71.61 ± 0.51    & 85.28 ± 0.35               \\
FRN  (2021)            & ResNet12          & 66.45 ± 0.19    & 82.83 ± 0.13            & 71.16 ± 0.22    & 86.01 ± 0.15               \\
% MCL-Katz         & ResNet12          & 67.51 ± n/a           & 83.99 ± n/a              & 72.01 ± n/a      & 86.02 ± n/a                \\
MCL (2022)   & ResNet12          & 67.45 ± n/a          & 84.36 ± n/a              & 72.01 ± n/a      & 86.31 ± n/a                 \\
% MCL-Katz pyramid & ResNet12          & 67.85 ± n/a          & 84.47 ± n/a              & 72.13 ± n/a      & 86.32 ± n/a                 \\
DeepBDC (2022)     & ResNet12          & 67.34 ± 0.43    & 84.46 ± 0.28            & {72.34 ± 0.49}    & \textbf{87.31 ± 0.32}               \\ 
\hline
% pre-train        & ResNet12          & 68.625 ±         & 85.073±                 & -               & -                          \\
Ours    & ResNet12       & \textbf{69.71 ± 0.45}        & \textbf{86.09 ± 0.27}           & \textbf{73.02 ± 0.50}               &  {87.25 ± 0.34}                           \\
\hline
\end{tabularx}
\label{testacc1}
\end{table*}

\begin{table*}[thbp]
\centering
\caption{Few-shot classification accuracy (\%) with 95\% confidence intervals on CUB and CIFAR-FS.}
\begin{tabularx}{\textwidth}{>{\centering\arraybackslash}X>{\centering\arraybackslash}X>{\centering\arraybackslash}X>{\centering\arraybackslash}X>{\centering\arraybackslash}X>{\centering\arraybackslash}X>{\centering\arraybackslash}X}
\hline
\multirow{2}{*}{Method} & \multirow{2}{*}{Backbone} & \multicolumn{2}{c}{\textbf{CUB}} & \multicolumn{2}{c}{\textbf{CIFAR-FS}} \\ \cline{3-6} 
                        &                           & 1shot          & 5-shot          & 1-shot           & 5-shot          \\ \hline  
FEAT (2020)         & ResNet12          & 73.27 ± 0.22   & 85.77 ± 0.14   & -                 & -                 \\
DeepEMD(2020)   & ResNet12          & 75.65 ± 0.83    & 88.69 ± 0.50    & -                 & -                 \\
S2M2 (2020)          & ResNet34          & 72.92 ± 0.83    & 86.55 ± 0.51    & 62.77 ± 0.23      & 75.75 ± 0.13      \\
CovNet (2020)        & ResNet18          & 80.76 ± 0.42      & 92.05 ± 0.20      & -                 & -                 \\
% InvEq distill \cite{rizve2021exploring}    & ResNet12          & -               & -               & {77.87 ± 0.85}      & \textbf{89.74 ± 0.57}      \\
ReNet (2021)          & ResNet12          & 79.49 ± 0.44    & 91.11 ± 0.24    & 74.51 ± 0.46      & 86.60 ± 0.32      \\
FRN (2021          & ResNet12          & 83.55 ± 0.19    & 92.92 ± 0.10      & -                 & -                 \\
% MCL pyramid fcn  & ResNet12          & 83.64 ± n/a     & 92.18 ± n/a     & -                 & -                 \\
MCL (2022)& ResNet12          & {83.64 ± n/a}     & 92.18 ± n/a     & -                 & -                 \\
DeepBDC (2022)   & ResNet12          & 83.55 ± 0.40    & {93.82 ± 0.17}    & -                 & -                 \\
TDM (2022)            & ResNet12          & 84.36 ±   0.19  & 93.37 ± 0.10    & -                 & -                 \\ \hline
% pre-train        & ResNet12          & 75.76           & 91.564          & 75.941            & 88.687            \\
Ours  & ResNet12          & \textbf{85.85 ± 0.42}          & \textbf{94.74 ± 0.20}          & \textbf{78.92 ± 0.46}            & \textbf{89.50 ± 0.30}            \\ \hline
\end{tabularx}
\label{testacc2}
\end{table*}

\subsection{Experimental Settings}
\subsubsection{Datasets}
Our meta-training process is evaluated on four popular benchmarks, namely miniImageNet, tieredImageNet, CUB, and CIFAR-FS. The miniImageNet dataset comprises 60,000 images evenly distributed over 100 object classes and is a subset of ImageNet \cite{russakovsky2015imagenet}. The dataset is split into training, validation, and testing sets, consisting of 64, 16, and 20 object classes, respectively. The tieredImageNet dataset is a challenging benchmark where the train, validation, and test sets are separated in terms of super-classes from the ImageNet hierarchy, necessitating better generalization than other datasets. The train, validation, and test sets consist of 20, 6, and 8 super-classes, respectively, and these super-classes comprise 351, 97, and 160 sub-classes. CUB-200-2011 (CUB) is a fine-grained classification dataset for bird species, containing 100/50/50 object classes for the train/validation/test splits, respectively. We follow recent work \cite{ye2020few,zhang2020deepemd,kang2021relational} and use pre-cropped images for human-annotated bounding boxes. Finally, CIFAR-FS is a dataset based on CIFAR-100 \cite{krizhevsky2009learning}, with 64, 16, and 20 object classes in the training, validation, and testing sets, respectively.
\vspace{-0.1cm}

\subsubsection{Pre-training} We select ResNet12 as the backbone network. The input image has a spatial size of 84 $\times$84, and the output of the backbone network is a 640-dimensional representation vector. The model is trained to simultaneously perform the whole-classification as well as the rotation-prediction tasks similar to \cite{2019Boosting,chen2021pareto,bendou2022easy}. We select SGD as the optimizer with a momentum of 0.9 and a weight decay of $5e^{-4}$. The learning rate starts at 0.1 and is gradually annealed to 0.
\vspace{-0.1cm}

\subsubsection{Meta-training} We use SGD as the optimizer during meta-training, with a momentum of 0.9 and a weight decay of $5e^{-4}$. We do not use any popular meta-training learning rate schedulers \cite{Wertheimer2021, Liu2021}, and fix the learning rate to $1e^{-3}$ for all experiments. Following \cite{Wertheimer2021}, which employs a larger way during meta-training, we train models with 20 ways.

\subsubsection{Evaluation} We evaluate classification accuracy on the testing
datasets. Specifically, a set of $N$-way $K$-shot tasks are randomly sampled from novel classes. Each task is created by selecting $N$ novel classes, and then randomly selecting $K$ support and 15 query images per class. The classification accuracy is measured on the query images for each task and we report average classification accuracy with 95\% confidence intervals of randomly sampled 2,000 test episodes.
\vspace{-0.2cm}

\subsection{Comparisons to State-of-the-art Methods}
To evaluate the effectiveness of our proposed methods, we conducted meta-training on benchmark datasets and present the results in Tables \ref{testacc1} and \ref{testacc2}. Our meta-trained model, utilizing both SKL and NNSKL, shows comparable performance on all datasets. Notably, our model even outperforms dense feature-based few-shot methods \cite{wertheimer2021few, Liu2021}, despite only utilizing global features from the backbone for inference. However, we must note that DeepBDC has two versions: Meta DeepBDC and STL DeepBDC. For comparison purposes, we only present our results for Meta DeepBDC, as STL DeepBDC utilizes extra data during the training phase.

\begin{table*}[th]
\caption{Ablative analysis of SKL and NNSKL. Results are few-shot classification accuracy (\%).  }
\begin{tabularx}{\textwidth}{>{\centering\arraybackslash}X>{\centering\arraybackslash}X>{\centering\arraybackslash}X>{\centering\arraybackslash}X>{\centering\arraybackslash}X>{\centering\arraybackslash}X>{\centering\arraybackslash}X>{\centering\arraybackslash}X>{\centering\arraybackslash}X}
\hline
\multirow{2}{*}{Model} & \multicolumn{2}{c}{\textbf{miniImageNet}}     & \multicolumn{2}{c}{\textbf{tieredImageNet}}     & \multicolumn{2}{c}{\textbf{CUB}}              & \multicolumn{2}{c}{\textbf{CIFAR-FS}}         \\ \cline{2-9} 
                       & 1-shot                & 5-shot                & 1-shot                & 5-shot                  & 1-shot                & 5-shot                & 1-shot                & 5-shot                \\ \hline
Pre-train              & 68.625          & 85.07           & 71.729         & 87.021          & 75.76         & 91.564         & 75.94           & 88.69          \\
SC                     & 68.77           & 84.75          & 71.88          & 86.06            & 85.02           & 94.07           & 77.99           & 88.58          \\
SC+SKL                 & 69.52           & 85.01          & 72.565          & 86.08            & 85.48           & 94.07           & 78.59          & 88.89          \\
SC+NNSKL               & 69.21           & \textbf{86.09 } & 72.89           & \textbf{87.25 } & 85.72          & 94.14           & 78.76          & 89.42          \\
SC+SKL+NNSKL           & \textbf{69.71 } & 85.71           & \textbf{73.02 } & 87.12           & \textbf{85.85} & \textbf{94.74 } & \textbf{78.92 } & \textbf{89.50} \\ \hline
\end{tabularx}
\label{tab:ablative}
\end{table*}

\begin{figure*}[th]
  \centering
    \includegraphics[width=\linewidth]{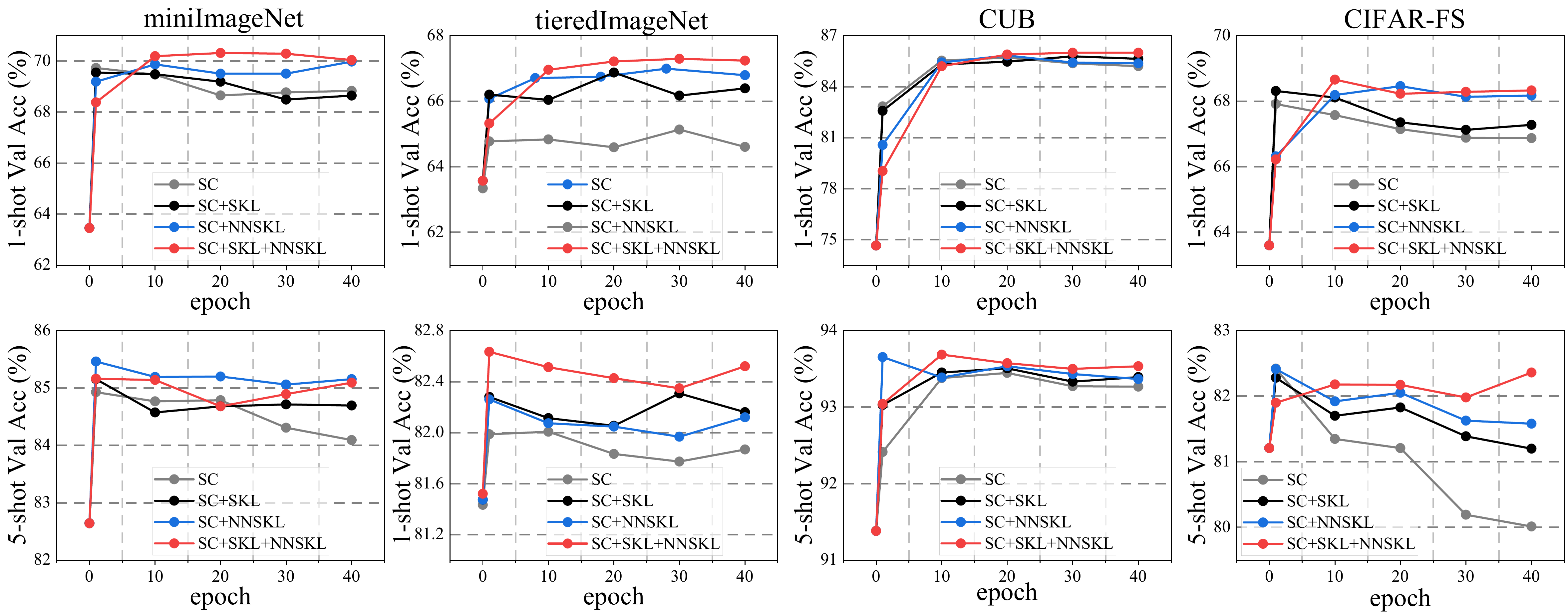}
\caption{Validation accuracy of meta-training processes on four benchmark datasets. The upper row shows 1-shot validation accuracy while the lower row reports 5-shot validation accuracy.}
  \label{fig:valcifar}
\end{figure*}

\subsection{Ablation Study}
\subsubsection{SKL and NNSKL}
To demonstrate the effectiveness of knowledge distillation techniques for few-shot learning, we conducted ablative studies on SKL and NNSKL and compared their performance to the model trained solely with supervised contrastive loss. As can be seen from Table \ref{tab:ablative}, knowledge-distillation techniques leads to better performance on all benchmark datasets. While single contrastive loss hinders the performance of 1-shot and 5-shot tasks of CIFAR-FS and 5-shot tasks of miniImageNet, combining supervised contrastive loss with knowledge distillation techniques improves the model's performance by distilling generalizable knowledge from the teacher model.  For each task, the model trained with both SKL and NNSKL achieved the best performance. Notably, knowledge distillation techniques marginally improved performance on the CUB dataset for both 1-shot and 5-shot tasks. This is due to the fine-grained nature of the CUB, which exhibits fewer gaps between the training and test datasets, leading to less over-discrimination during meta-training. For this type of dataset, the base class discrimination is enough to boost performance without worrying about over-discrimination. 

Moreover, we present the validation accuracy during meta-training on the CIFAR-FS dataset. As illustrated in Figure \ref{fig:valcifar}, the knowledge distillation techniques result in higher validation accuracy throughout the meta-training process, demonstrating the efficacy of SKL and NNSKL. NNSKL outperforms SKL on the miniImageNet and CIFAR-FS datasets, while SKL performs slightly better on tieredImageNet, and they have similar performance on CUB. Using both SKL and NNSKL yields the highest validation accuracy on all tasks except the 5-shot task on miniImageNet. The task validation accuracies on CUB remained stable after reaching the plateau, demonstrating less over-discrimination on the CUB.

% \textbf{Effect of frozen linear classifier}
% To understand the effect of the frozen linear classifier, we compare the performance between models optimizing or freezing the linear classifier. We select the tasks on CIFAR-FS, as it suffers the most from over-discrimination. 

\begin{figure}[th]
  \centering
    \includegraphics[width=0.99\linewidth]{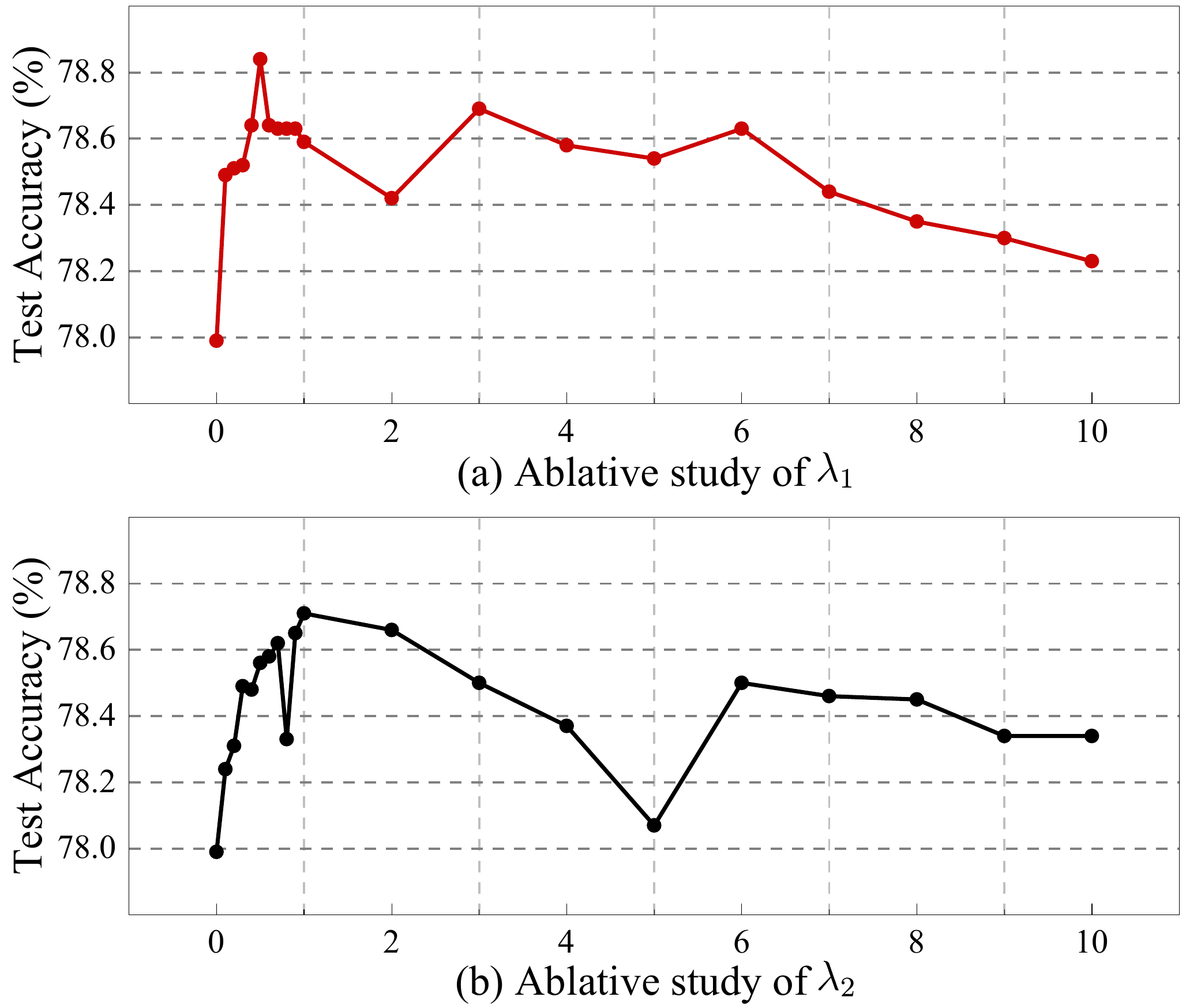}
\caption{Ablative study of $\lambda_1$ and $\lambda_2$. When $\lambda_1=0$ or $\lambda_2=0$, the model is trained with only supervised contrastive loss.}
  \label{fig:ablative}
\end{figure}

\subsubsection{Hyperparameters}
Then, we studied the influence of different auxiliary weights $\lambda_1$ and $\lambda_2$. Figure \ref{fig:ablative} reports the ablative results of $\lambda_1$ and $\lambda_2$ on the CIFAR-FS dataset, respectively. For fair comparisons, $\lambda_2$ ($\lambda_1$) is set to 0 when ablating $\lambda_1$ ($\lambda_2$). As can be seen in Figure \ref{fig:ablative} (a), the performance arises along with the increase of $\lambda_1$ and then drops due to the stronger restriction of the base class discrimination. The same trend can be observed in Figure \ref{fig:ablative} (b). It is worth noting that, though the performance varies, models trained with our meta-training process consistently outperform the standard meta-trained model ($\lambda_1=\lambda_2=0$). 

% \begin{table*}[thbp]
% \label{tab:lambda}
% \caption{Ablation studies on the auxiliary weight $\lambda$ on 5-way-1-shot performance on CIFAR-FS with NNSKL.}
% \begin{tabularx}{\textwidth}{>{\centering\arraybackslash}X>{\centering\arraybackslash}X>{\centering\arraybackslash}X>{\centering\arraybackslash}X>{\centering\arraybackslash}X>{\centering\arraybackslash}X>{\centering\arraybackslash}X>{\centering\arraybackslash}X>{\centering\arraybackslash}X>{\centering\arraybackslash}X>{\centering\arraybackslash}X>{\centering\arraybackslash}X>{\centering\arraybackslash}X}
% \hline
% $\lambda$ & 0 & 1 & 2 & 3 & 4 & 5 & 6 & 7 & 8 & 9 & 10 \\ \hline
% 1-shot    &   &     &     &     &     &     &     &     &     &     &   &   &   &   &   &   &   &   &   &    \\ \hline
% 5-shot    &   &     &     &     &     &     &     &     &     &     &   &   &   &   &   &   &   &   &   &    \\ \hline
% \end{tabularx}
% \end{table*}

\subsection{Analysis}
% \noindent\textbf{Decreased Base Class Discrimination.}
% To demonstrate the effectiveness of SKL and NNSKL, we conducted singular value decomposition on embedding matrices on the training set of CIFAR-FS (see Figure \ref{fig:collapse}). We observe that supervised contrastive loss causes severe dimensional collapse. and both SKL and NNSKL mitigate dimension collapse, thus mitigating base class discrimination. We also observe that NNSKL has the best performance, as it results in the least collapsed spectrum.

% \noindent\textbf{Increased Novel Class Generalization.} 

\begin{figure}[t]
  \centering
    \includegraphics[width=\linewidth]{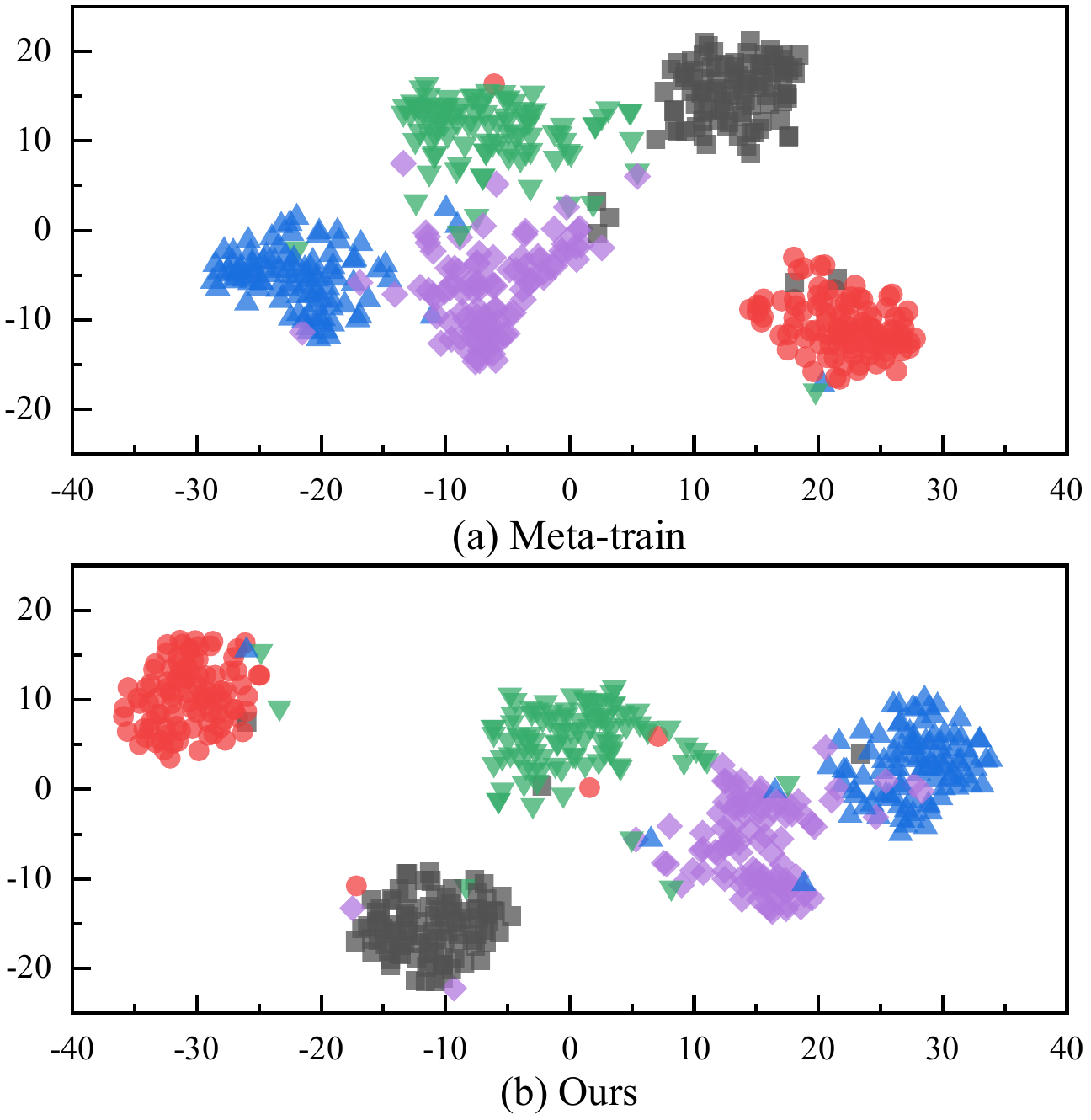}
\caption{T-SNE visualization of the feature distribution of CIFAR-FS. Different colors correspond to different classes and 100 embeddings are sampled from each novel class.}
  \label{fig:tsne}
\end{figure}

In this paper, we also conducted a T-SNE visualization on the embeddings extracted by the backbone network to demonstrate the effectiveness of the proposed framework in handling novel classes of CIFAR-FS, as presented in Figure  \ref{fig:tsne}. The results indicate that the standard meta-training method leads to smaller inter-class distances and higher intra-class variances compared to our proposed meta-training process, resulting in poor generalization ability towards novel tasks. The incorporation of either the SKL or NNSKL loss function in meta-training is found to decrease the intra-class variance while preserving the inter-class distance, leading to better generalization performance towards novel classes.

\begin{figure}[t]
  \centering
    \includegraphics[width=\linewidth]{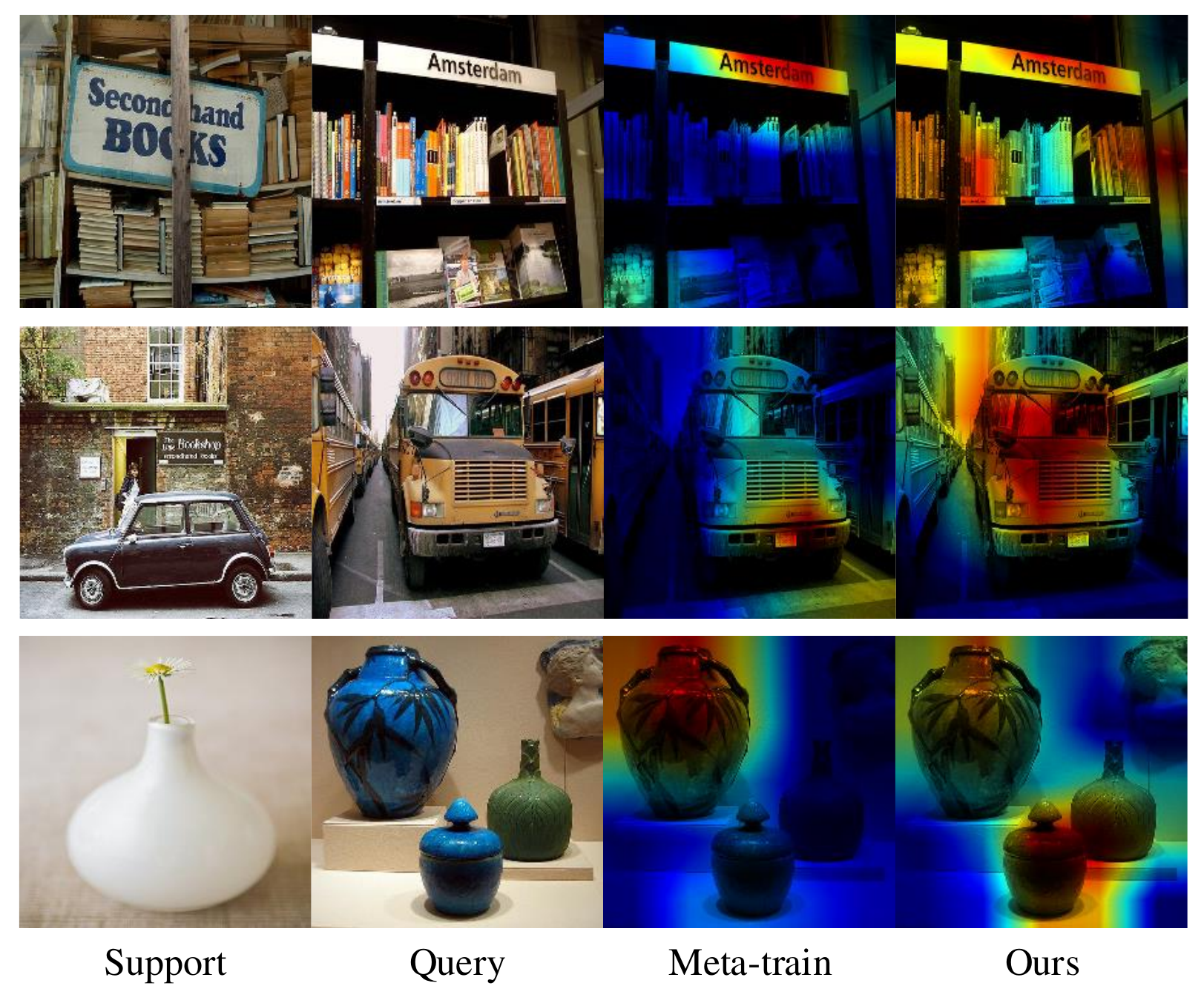}
\caption{Channel attention map of query images with respect to the support images.}
  \label{fig:cam}
\end{figure}

Moreover, we present the channel attention map of the query feature map with respect to the globally averaged feature of the corresponding support image in Figure \ref{fig:cam}. The images used for this analysis are randomly sampled from the miniImageNet test set, and attention maps are bi-linearly interpolated to the input image size. Results reveal that the standard meta-training method over-emphasizes the discriminative features while disregarding the generalizable ones, as shown in the attention maps on bottles and books. In contrast, our proposed meta-training process is found to maintain the generalization ability, thereby leading to better performance in handling novel classes.

\section{Conclusion}
This study has identified that over-discrimination causes overfitting in the meta-training process. To address this issue, the standard symmetric KL divergence (SKL) has been introduced, which takes a batch of images as input and restricts the output class distribution of the linear classifier of the student model to be similar to that of the teacher model. SKL preserves the generalizable knowledge by implicitly limiting the movement of the student embeddings from the teacher embeddings. To further combat over-discrimination for few-shot classification, the Nearest Neighbor Symmetric KL divergence (NNSKL) was proposed. NNSKL takes a batch of few-shot tasks as input and penalizes the output distribution of the nearest neighbor classifier. This approach targets the relationships between query embedding and support class centers, which have a stronger impact on the deep metric space. Additionally, since the supervised contrastive loss also consists of the output distribution of the nearest neighbor classifier, it can be seamlessly integrated into the meta-training process. By combining SKL and NNSKL with the supervised contrastive loss, the proposed meta-training process considers novel class generalization during training and consistently improved few-shot performance on popular benchmark datasets.

\bibliographystyle{ACM-Reference-Format}
\bibliography{main}
\end{document}

\title{Supplementary Material: Understanding the Overfitting of the Episodic Meta-training}





\maketitle

\section{Decreased Dimension Collapse} 
\begin{figure}[h]
  \centering
    \includegraphics[width=1\linewidth]{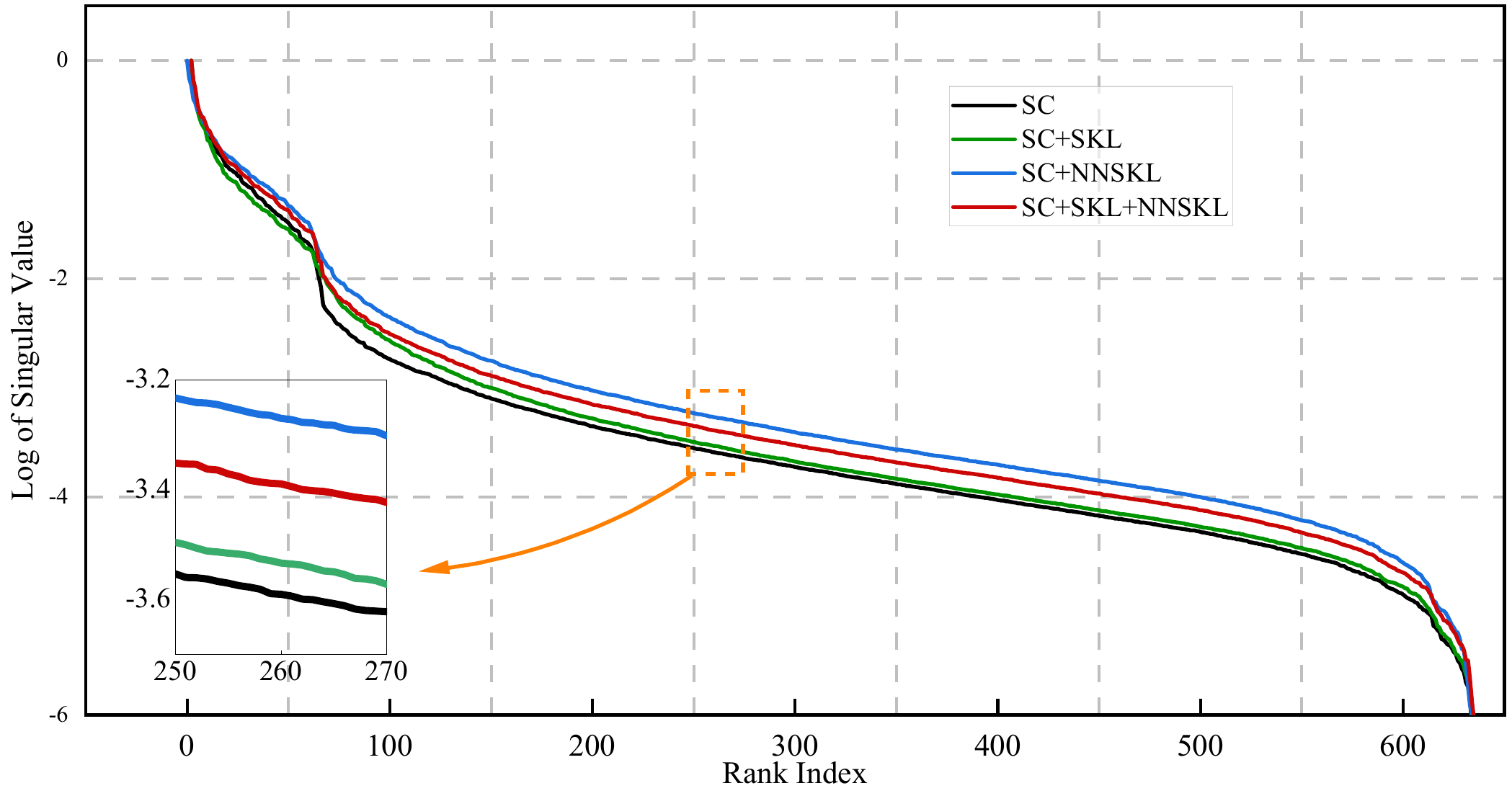}
  \caption{Singular value spectrums of embedding matrices on the training set of miniImageNet. Each spectrum contains 640 singular values in sorted order and logarithmic scale. SC is the model trained with only supervised contrastive loss, SKL or NNSKL indicates the model is assisted by the SKL or NNSKL loss.}
  % \caption{The training loss curves (orange) and validation accuracy curves (blue) in the meta-training stage.}
  \label{fig:sig_mini}
  \vspace{2cm}
\end{figure}

\begin{figure}[thbp]
  \centering
    \includegraphics[width=1\linewidth]{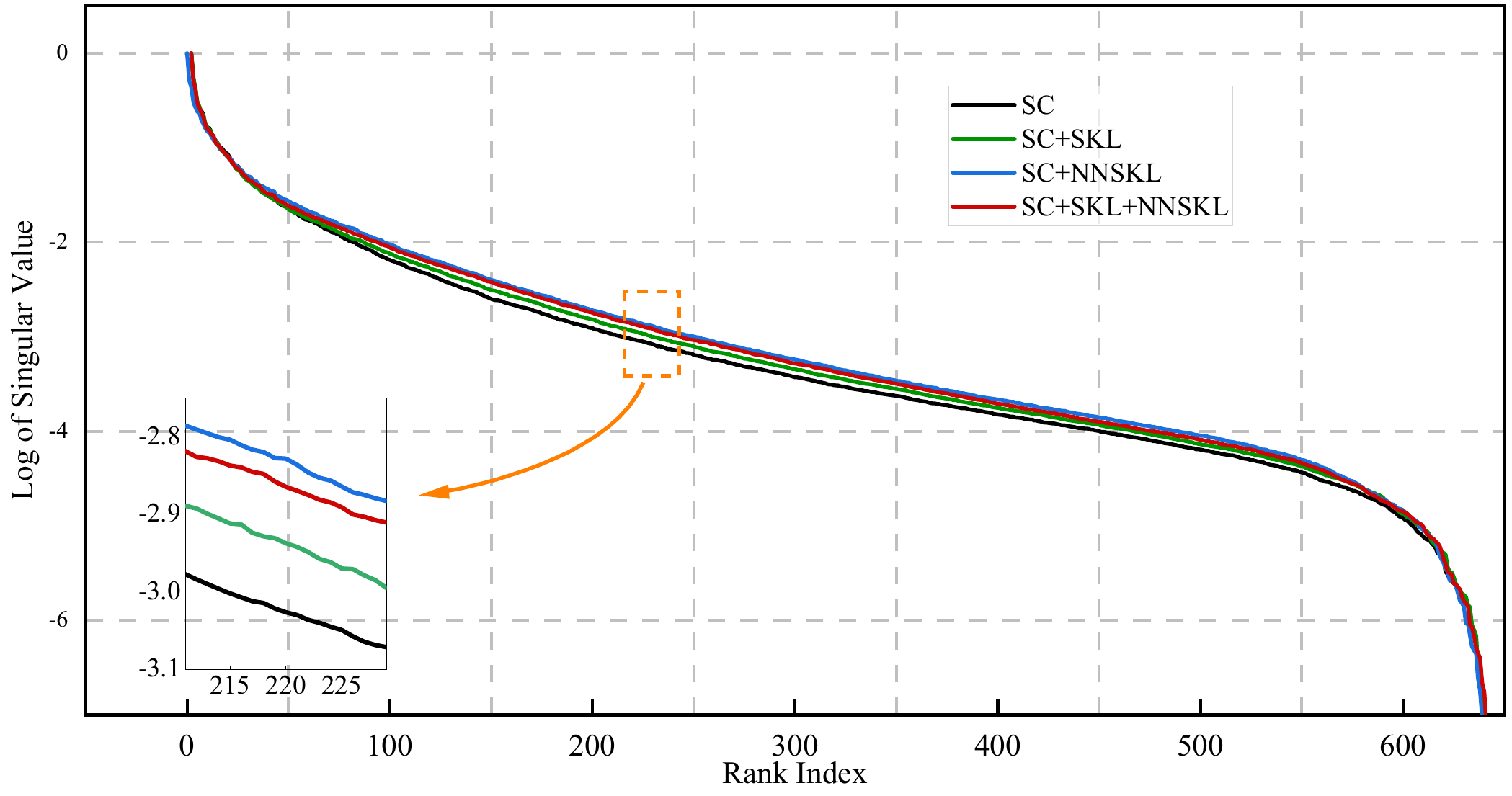}
  \caption{Singular value spectrums of embedding matrices on the training set of tieredImageNet. Each spectrum contains 640 singular values in sorted order and logarithmic scale. SC is the model trained with only supervised contrastive loss, SKL or NNSKL indicates the model is assisted by the SKL or NNSKL loss.}
  % \caption{The training loss curves (orange) and validation accuracy curves (blue) in the meta-training stage.}
  \label{fig:sig_tiered}
\end{figure}

\begin{figure}[htbp]
  \centering
    \includegraphics[width=1\linewidth]{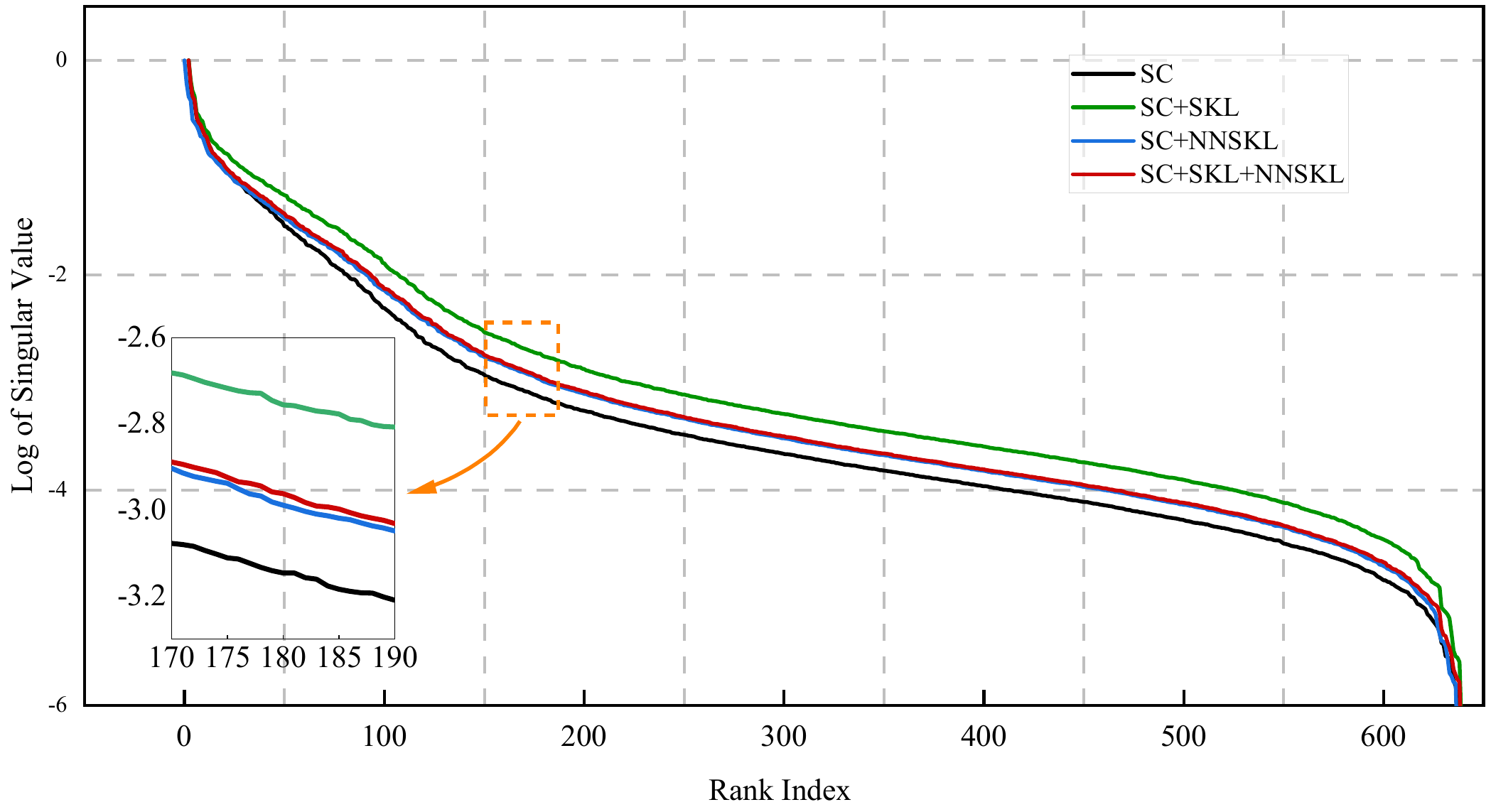}
  \caption{Singular value spectrums of embedding matrices on the training set of CUB. Each spectrum contains 640 singular values in sorted order and logarithmic scale. SC is the model trained with only supervised contrastive loss, SKL or NNSKL indicates the model is assisted by the SKL or NNSKL loss.}
  % \caption{The training loss curves (orange) and validation accuracy curves (blue) in the meta-training stage.}
  \label{fig:sig_cub}
\end{figure}

\begin{figure}[ht]
  \centering
    \includegraphics[width=1\linewidth]{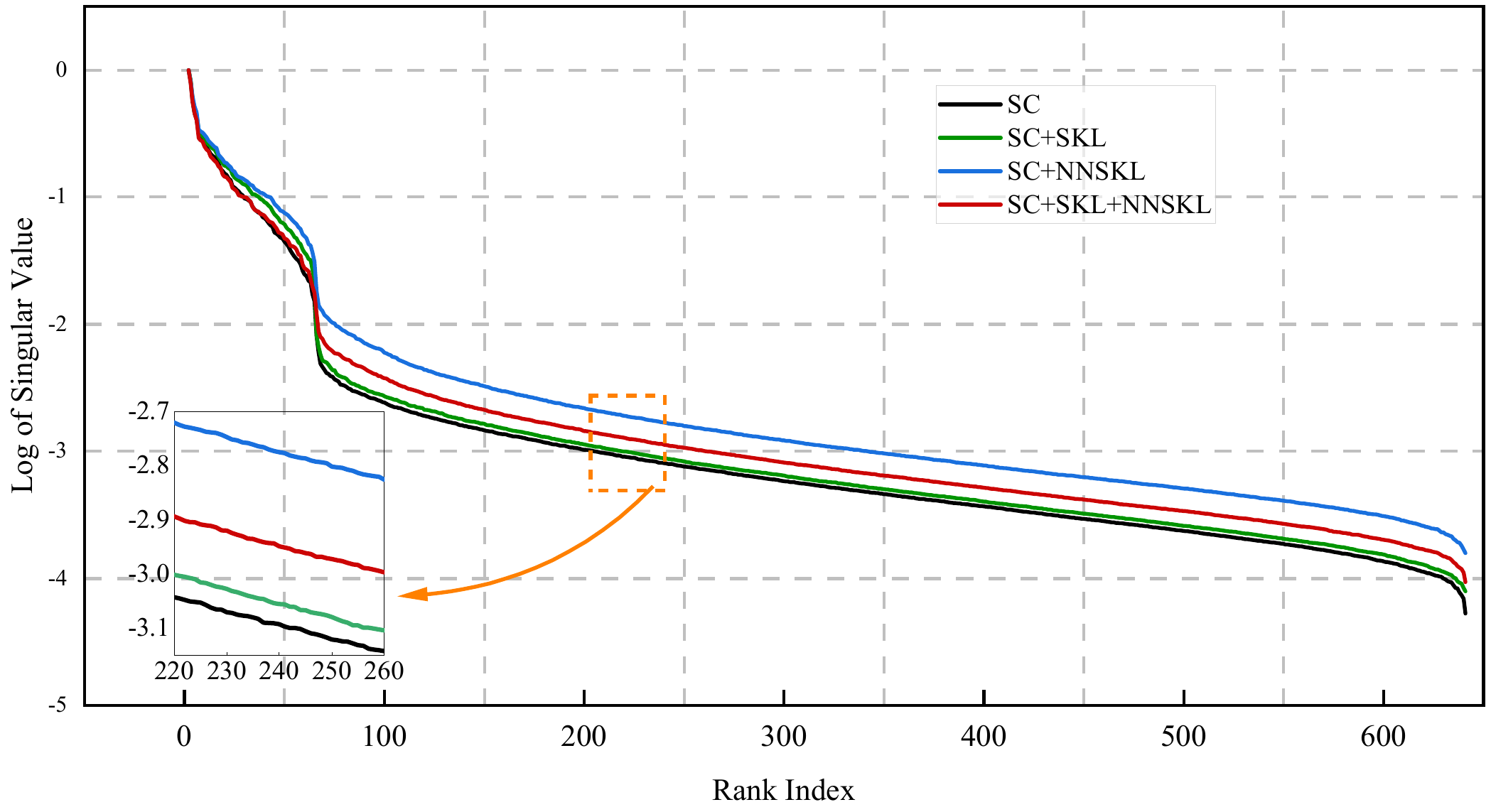}
  \caption{Singular value spectrums of embedding matrices on the training set of CIFAR-FS. Each spectrum contains 640 singular values in sorted order and logarithmic scale. SC is the model trained with only supervised contrastive loss, SKL or NNSKL indicates the model is assisted by the SKL or NNSKL loss.}
  % \caption{The training loss curves (orange) and validation accuracy curves (blue) in the meta-training stage.}
  \label{fig:sig_cifar}
\end{figure}

The effectiveness of SKL and NNSKL in mitigating the problem of over-discrimination and preventing dimension collapse is demonstrated by analyzing the spectra of the embedding matrices of the training sets of benchmark datasets. To this end, we extracted embedding matrices from models that were trained for 40 epochs to perform 5-shot tasks and conducted singular value decompositions on them. The resulting spectrums for miniImageNet, tieredImageNet, CUB, and CIFAR-FS are depicted in Figure \ref{fig:sig_mini} to \ref{fig:sig_cifar}. The analysis of the spectrums reveals that both SKL and NNSKL are effective in preventing dimension collapse, with models trained using either method exhibiting higher eigenvalue counts for a larger range of dimensions. Interestingly, SKL performs best on CUB, while NNSKL outperforms SKL on the remaining datasets in terms of dimension collapse.

\section{More Visual Results}
In this section, we provide more visual results, including T-SNE visualization and channel activation map, to illustrate the over-discrimination and the effect of SKL and NNSKL.

\subsection{T-SNE Visualization}
\begin{figure}[thbp]
  \centering
    \includegraphics[width=0.8\linewidth]{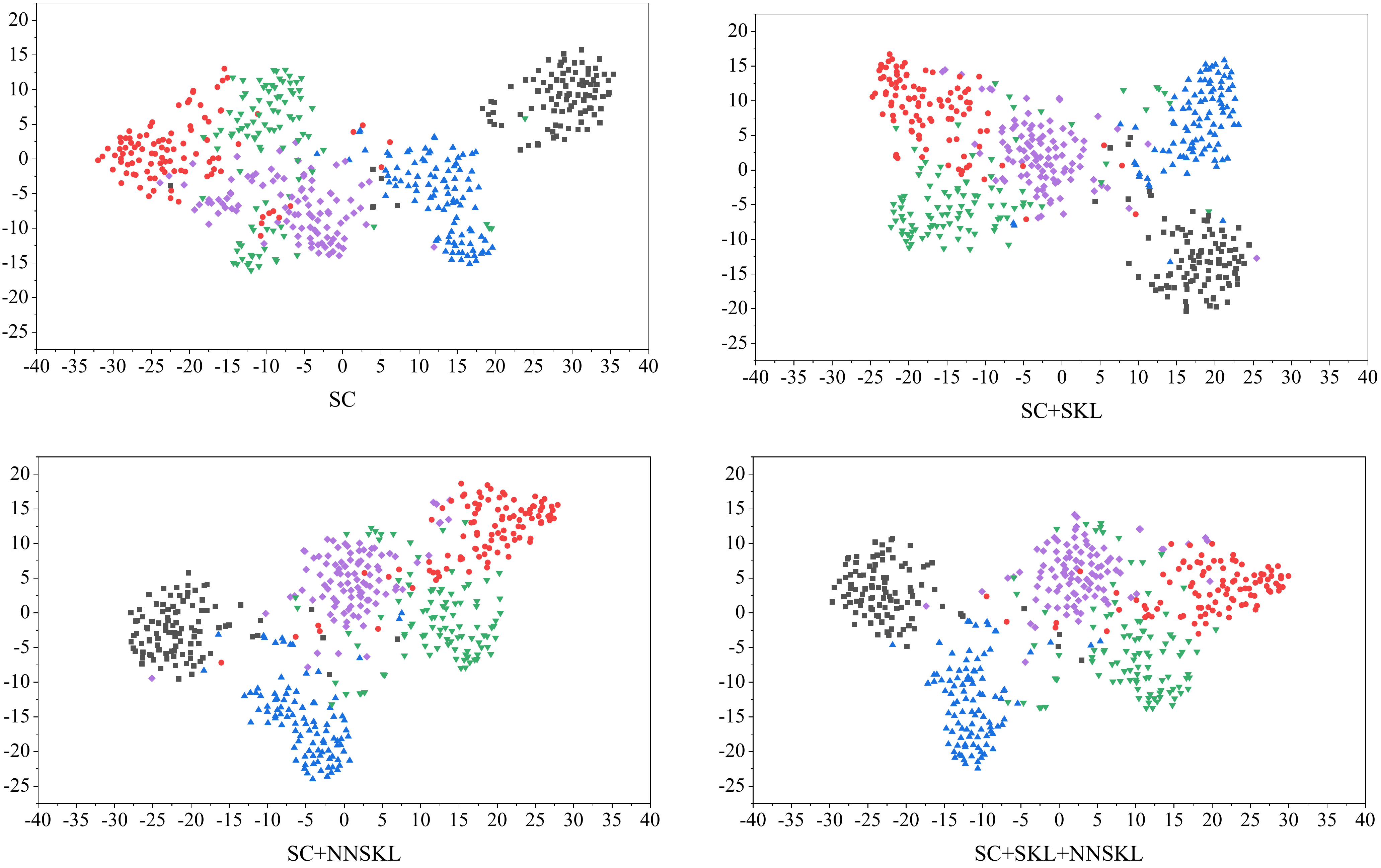}
\caption{T-SNE visualization of the feature distribution osf miniImageNet. Different colors correspond to different classes and 100 embeddings are sampled from each novel class. SC is the model trained with only supervised contrastive loss, SKL or NNSKL indicates the model is assisted by the SKL or NNSKL loss.}
  \label{fig:tsne_mini}
\end{figure}

\begin{figure}[thbp]
  \centering
    \includegraphics[width=0.8\linewidth]{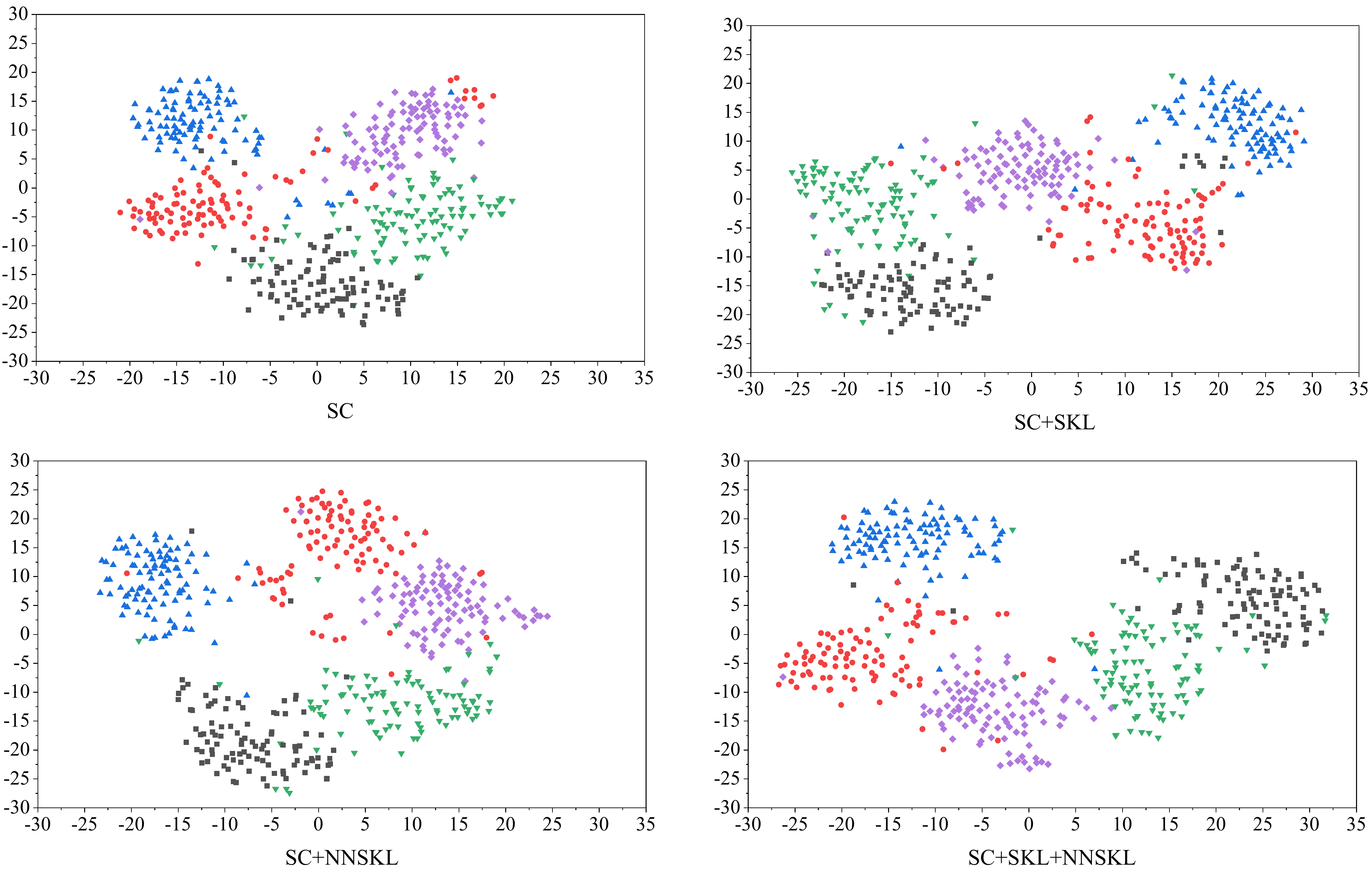}
\caption{T-SNE visualization of the feature distribution of tieredImageNet. Different colors correspond to different classes and 100 embeddings are sampled from each novel class. SC is the model trained with only supervised contrastive loss, SKL or NNSKL indicates the model is assisted by the SKL or NNSKL loss.}
  \label{fig:tsne_tiered}
\end{figure}

\begin{figure}[thbp]
  \centering
    \includegraphics[width=0.8\linewidth]{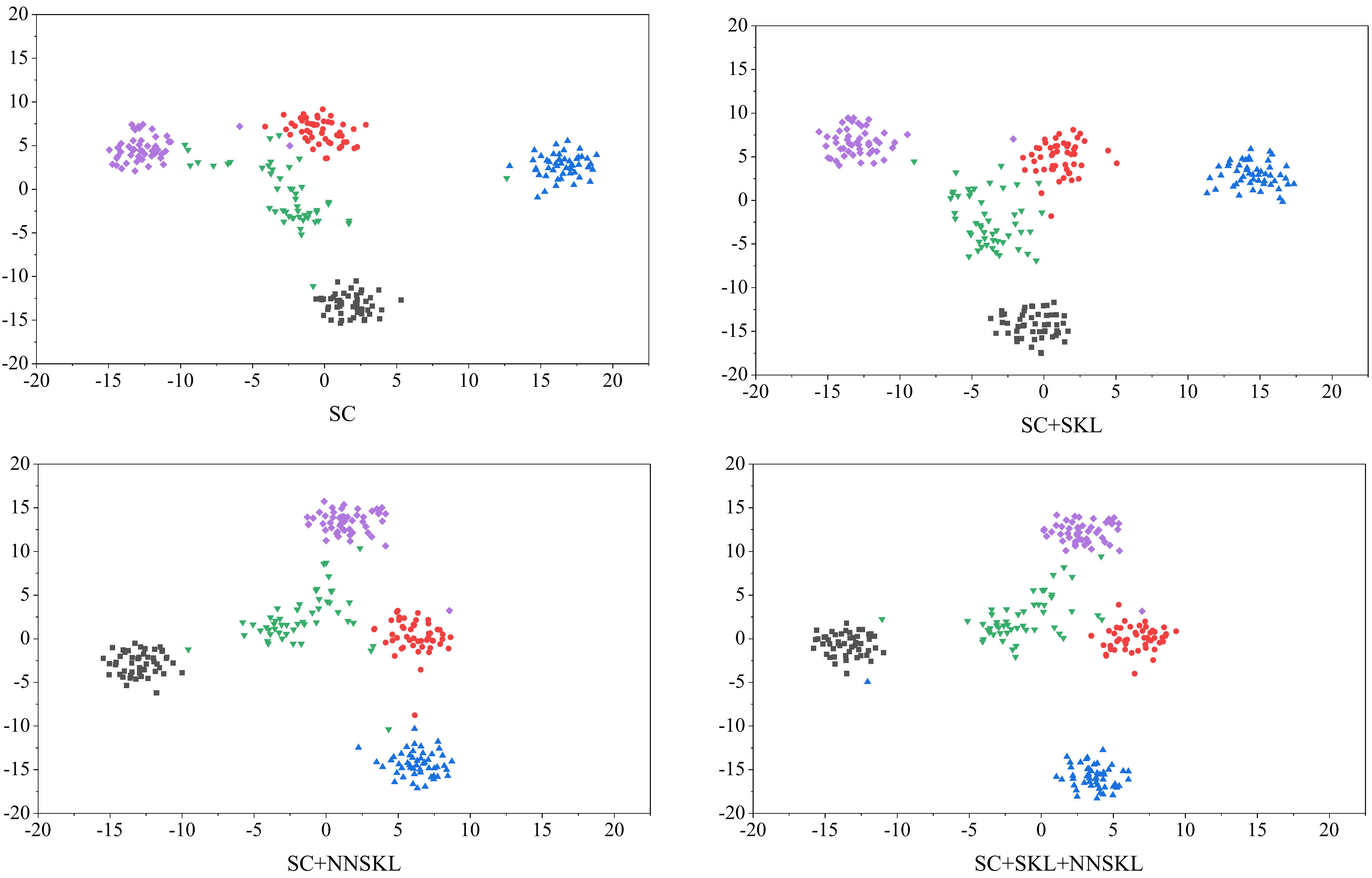}
\caption{T-SNE visualization of the feature distribution of CUB. Different colors correspond to different classes and 50 embeddings are sampled from each novel class. SC is the model trained with only supervised contrastive loss, SKL or NNSKL indicates the model is assisted by the SKL or NNSKL loss.}
  \label{fig:tsne_cub}
\end{figure}

\begin{figure}[thbp]
  \centering
    \includegraphics[width=0.8\linewidth]{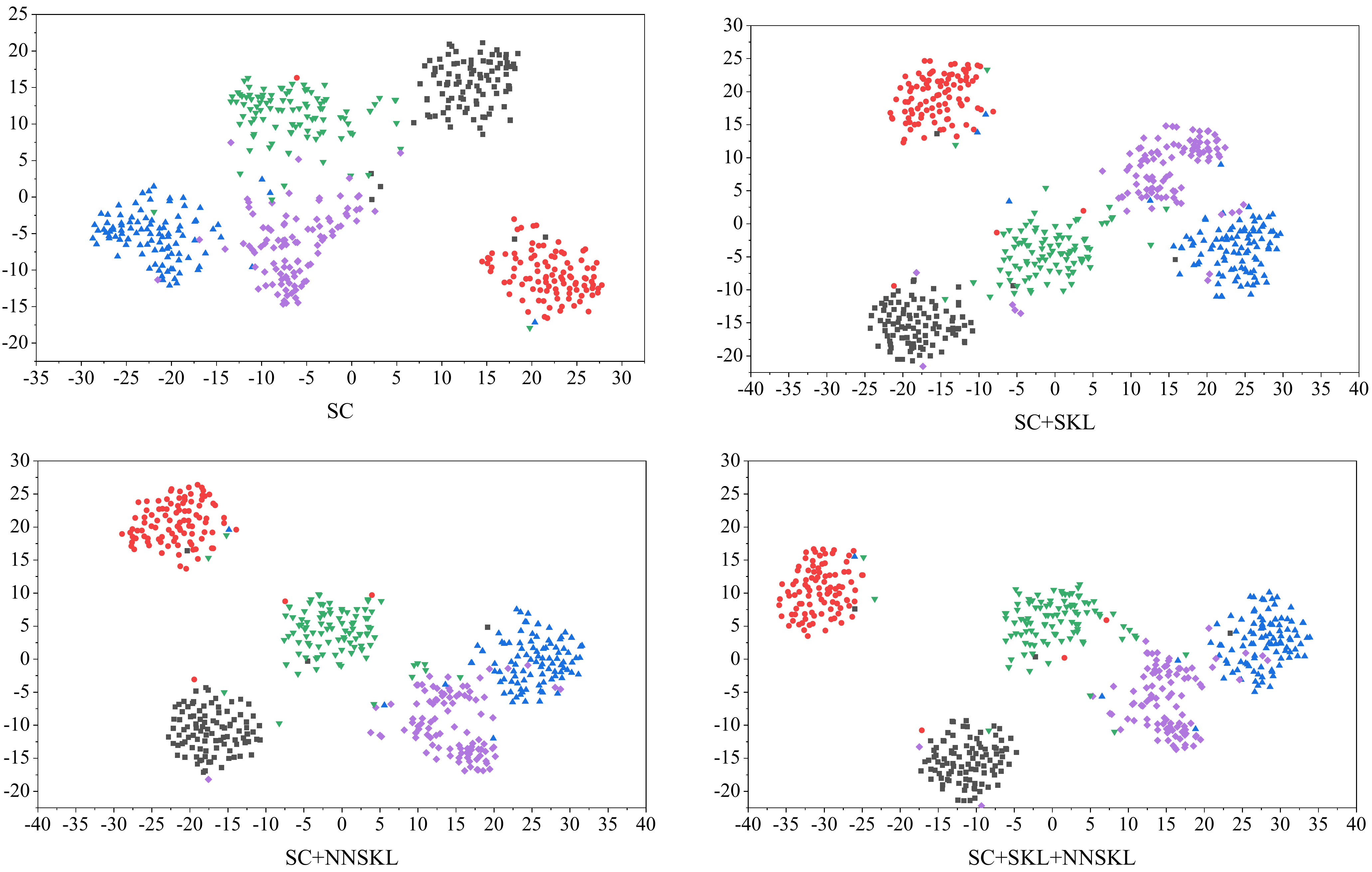}
\caption{T-SNE visualization of the feature distribution of CIFAR-FS. Different colors correspond to different classes and 100 embeddings are sampled from each novel class. SC is the model trained with only supervised contrastive loss, SKL or NNSKL indicates the model is assisted by the SKL or NNSKL loss.}
  \label{fig:tsne_cifar}
\end{figure}

To gain a comprehensive understanding of the impact of SKL and NNSKL on the deep metric space, we present t-SNE visualizations of the feature distributions on three different datasets, namely miniImageNet, tieredImageNet, CUB and CIFAR-FS, in Figure \ref{fig:tsne_mini} to \ref{fig:tsne_cifar}, respectively. To generate these visualizations, we first select five novel classes from each dataset and randomly sample 100 images for each class, except for CUB, where we sample 50 images. We then extract embeddings from these images and use them to perform t-SNE visualization. The models used for feature extraction are trained for 40 epochs to perform the 5-shot task on each dataset. These visualizations enable us to gain insights into the structure of the feature space and the impact of SKL and NNSKL on it.
\vspace{3.0cm}

\subsection{Channel Activation Map}
\begin{figure}[thbp]
  \centering
    \includegraphics[width=0.9\linewidth]{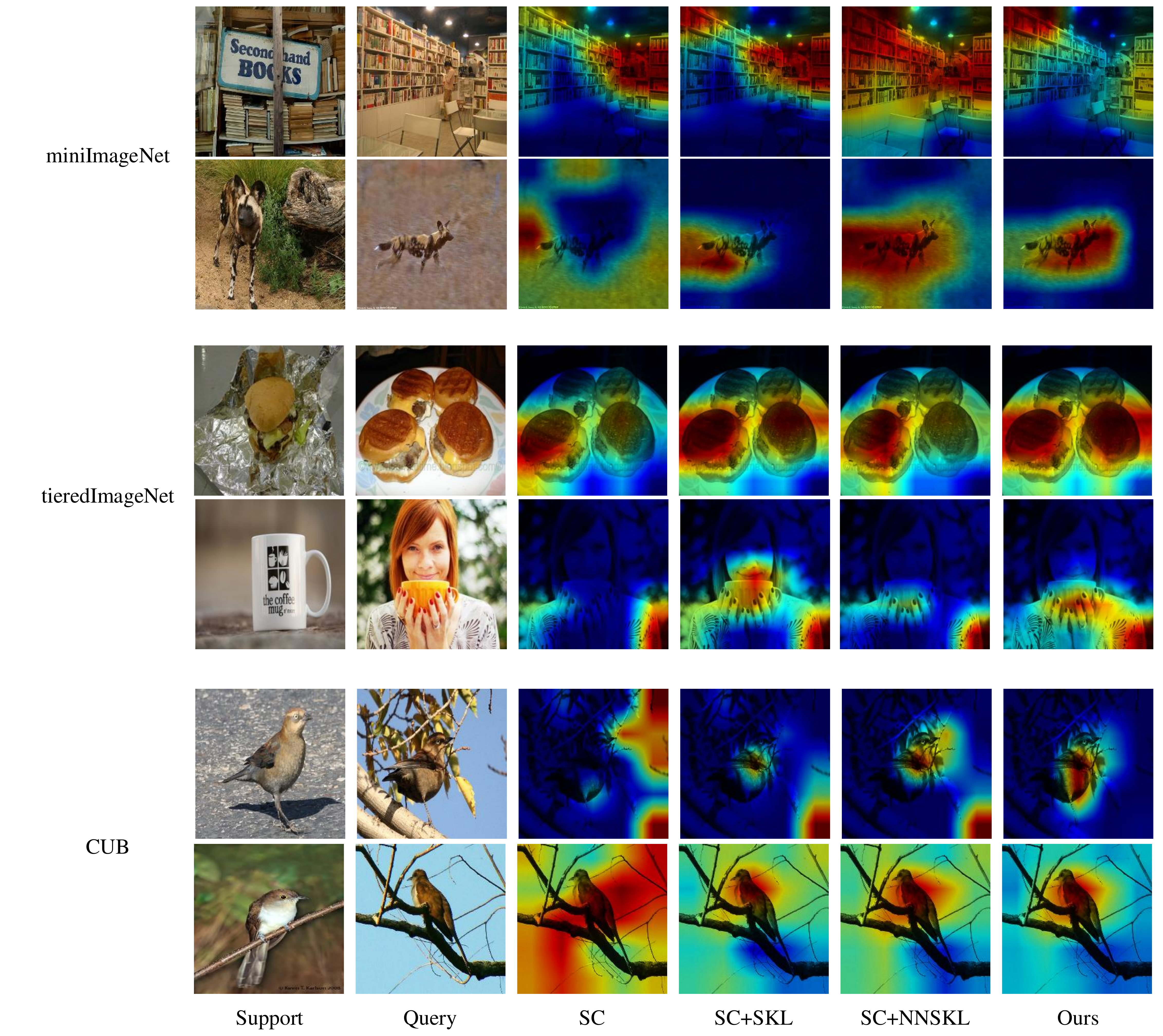}
\caption{Samples of channel activation maps of different models. SC is the model trained with only supervised loss. SKL (NNSKL) indicates that the model is assisted by the SKL (NNSKL). Ours indicates the model is trained with our meta-training process.}
  \label{fig:cam}
\end{figure}

In this section, we provide an intuitive understanding of the impact of SKL and NNSKL on the generalization of novel classes by examining the channel activation maps (CAM) on various datasets. As shown in Figure \ref{fig:cam}, the standard meta-trained model sometimes fails to concentrate on class-relevant regions, while knowledge-distillation-based models successfully pay attention to class-relevant areas. The presented CAMs were generated by models trained using different meta-training processes for performing the 1-shot task. The CAMs were created by computing the cosine correlation between query features of the feature map before global average pooling and the corresponding global-averaged support embedding. To ensure consistency in image size, the CAMs were bilinearly interpolated to match the size of the input image. The use of CAMs allows for the visualization of the model's attention on different regions of the input image, thereby providing an effective tool for evaluating the model's generalization ability. Notably, as the image resolution of the CIFAR-FS is too small (32 $\times$ 32), we do not show CAMs of the CIFAR-FS.

